\documentclass[10pt,letterpaper]{article}
\usepackage[a4paper,vmargin={20mm,20mm},hmargin={20mm,20mm}]{geometry}
\usepackage{apacite}
\usepackage{paralist}
\usepackage{tipa}
\usepackage[square,numbers,compress]{natbib}
\usepackage{threeparttable}
\usepackage{booktabs}
\usepackage{multirow}
\usepackage{url}
\usepackage[percent]{overpic}
\usepackage{bm}
\usepackage{float}
\usepackage[T1]{fontenc}
\usepackage{authblk}
\usepackage{amsmath}
\usepackage{amsfonts}
\usepackage[table]{xcolor}
\usepackage{tabularx}

\newcommand\eg{\textit{e.g.,~}}
\newcommand\ie{\textit{i.e.,~}}

\makeatletter
  \def\Hv@scale{.95}
\makeatother
\DeclareRobustCommand{\tablefont}{%
        \fontencoding{\encodingdefault}%
        \fontseries{m}
        \fontshape{n}
        \fontfamily{phv}
        \fontsize{8}{11}
        \selectfont}
\DeclareTextFontCommand{\texttable}{\tablefont}

\title{How Adults Understand What Young Children Say}

\author[a,b,\textdagger]{Stephan C. Meylan}
\author[c]{Ruthe Foushee} 
\author[a]{Nicole H. Wong} 
\author[b]{Elika Bergelson}
\author[a]{Roger P. Levy}

\renewcommand\Affilfont{\fontsize{9}{10.8}\itshape}

\affil[a]{Department of Brain and Cognitive Sciences, Massachusetts Institute of Technology}
\affil[b]{Department of Psychology and Neuroscience, Duke University}
\affil[c]{Department of Psychology, University of Chicago}

\begin{document}

\maketitle

\begin{abstract}

\noindent Children’s early speech often bears little resemblance to that of adults, and yet parents and other caregivers are able to interpret that speech and react accordingly.
Here we investigate how these adult inferences as listeners reflect sophisticated beliefs about what children are trying to communicate, as well as how children are likely to pronounce words.
Using a Bayesian framework for modeling spoken word recognition, we find that computational models can replicate adult interpretations of children's speech only when they include strong, context-specific prior expectations about the messages that children will want to communicate.
This points to a critical role of adult cognitive processes in supporting early communication and reveals how children can actively prompt adults to take actions on their behalf even when they have only a nascent understanding of the adult language.
We discuss the wide-ranging implications of the powerful listening capabilities of adults for theories of first language acquisition.


\begin{center}
\noindent \textbf{Keywords} 
\\
\vspace{1mm}
communication, child development, language development,   Bayesian inference,  psycholinguistics, speech recognition
\end{center}
\end{abstract}
\vspace{5mm}




How children learn their first language from the world around them has proven to be an enduring question in psychology, cognitive science, linguistics, and related disciplines \citep{chomsky1965, pinker1979, saffranEtAl1996, landauer1997solution, dupoux2018cognitive}. 
The degree to which adult caregivers influence the outcome of the language learning process has been central to research on this topic \citep{hoff2006, onnis2017, markusEtAl2000, RoseberryEtAl2014, rowlandEtAl2003, steinEtAl2008, fusaroli2021caregiver}. 
However, prior work has focused almost exclusively on adults' role as \textit{speakers} in child-centric interactions (\eg child-directed speech,  \citep{newport1975motherese, huttenlocher1991early, hart1995meaningful, rowe2012, golinkoffEtAl2019, cartmillEtAl2013, weizmanSnow2001,  bergelson2019north, cristia2019child}).
We argue here, however, that adults' role as speakers represents only one aspect of the ``work'' that adults put in to communication with young children: successful adult-child communication also depends on adults' highly sophisticated \textit{listening} behavior, which enables them to attribute word interpretations (and hence meanings) to children's variable and noisy linguistic output \citep{golinkoff1986beg, golinkoffGordon1988, tomasello1990young}. 
This allows adults to make sense of children's early language productions, which often deviate substantially from their own.
Understanding how adults interpret children's early speech is important because these interpretations help drive how caretakers respond and act, which in turn constitutes new input to the child in early language learning.


In this paper we ask what inferential process allows adults to interpret young children's highly variable speech.
First, what leads adults to determine that a child's vocalization\footnote{We focus here on the case of children producing spoken language, though the current proposal readily extends to signed languages.\\$~~~~~^{\dag}$ Corresponding author} is even interpretable, versus unintelligible?
Second, to what degree do the words interpreted by adults reflect the \textit{phonetic productions} produced by children, versus adult beliefs about what they expect children to say? 

Here we propose and test the hypothesis of \textit{child-directed listening}: in order to interpret the speech of young children, adult listeners must both rely heavily on their beliefs about what children are likely to say and
adapt to child pronunciations.
For example, if a child says ``ah wan du weed'' when a book is present, 
an adult will most likely take her to mean \textit{I want to read}, not \textit{I want to weed}. 
In such an example, adult beliefs about the child's communicative intent may override standard cues to word identity based on what the word \textit{sounds} like. 
That is, while the phonetic form [wid] (pronounced ``weed'') might typically suggest \textit{weed} as an adult speaker's intended word when judged solely based on acoustic information, an adult caregiver might instead assume that a \textit{child's} intended word was \textit{read}---both because children often pronounce \textit{r} as \textit{w}, and because children are more likely to want to \textit{read} than \textit{weed}.
In other cases, children may produce highly ambiguous vocalizations that adults interpret as words in the adult language by relying on cues from the communicative context (\eg interpreting ``da'' as \textit{dog} or \textit{dad} depending on context).
In both cases, adult caregivers use strong expectations about what children are likely to say in order to support sustained communication.
However, this idea has not been tested formally or at scale.

Though we envision that some version of child-directed listening is active across the early years of speech production, here we focus on adults' interpretations of child speech from when they are 12 months to 48 months old. 
The beginning of this interval corresponds to when adults first report that their children use words \citep{frank2021variability}.
We expect adults' beliefs to play the greatest role in informing their interpretation in these early stages when children's pronunciations differ the most from those of adults.
We use a collection of audio recordings of caregiver--child interactions (the Providence corpus; \citep{demuthEtAl2006, demuth2009prosodic}), and take transcribers' written annotations in the adult language as our proxy for caregivers' interpretations of what young children said.
We try to recover the word identities from the phonetic input and surrounding linguistic context using a suite of computational cognitive models of spoken word recognition.
We then compare the models' guesses to what adult transcribers \textit{actually} recovered from the children's speech. 
By parametrically varying properties of the models, we can infer the relative importance of different information sources---including the actual phonetic signal produced by children---for how adults understand young children.

\begin{figure*}
\centering
\includegraphics[width=\linewidth]{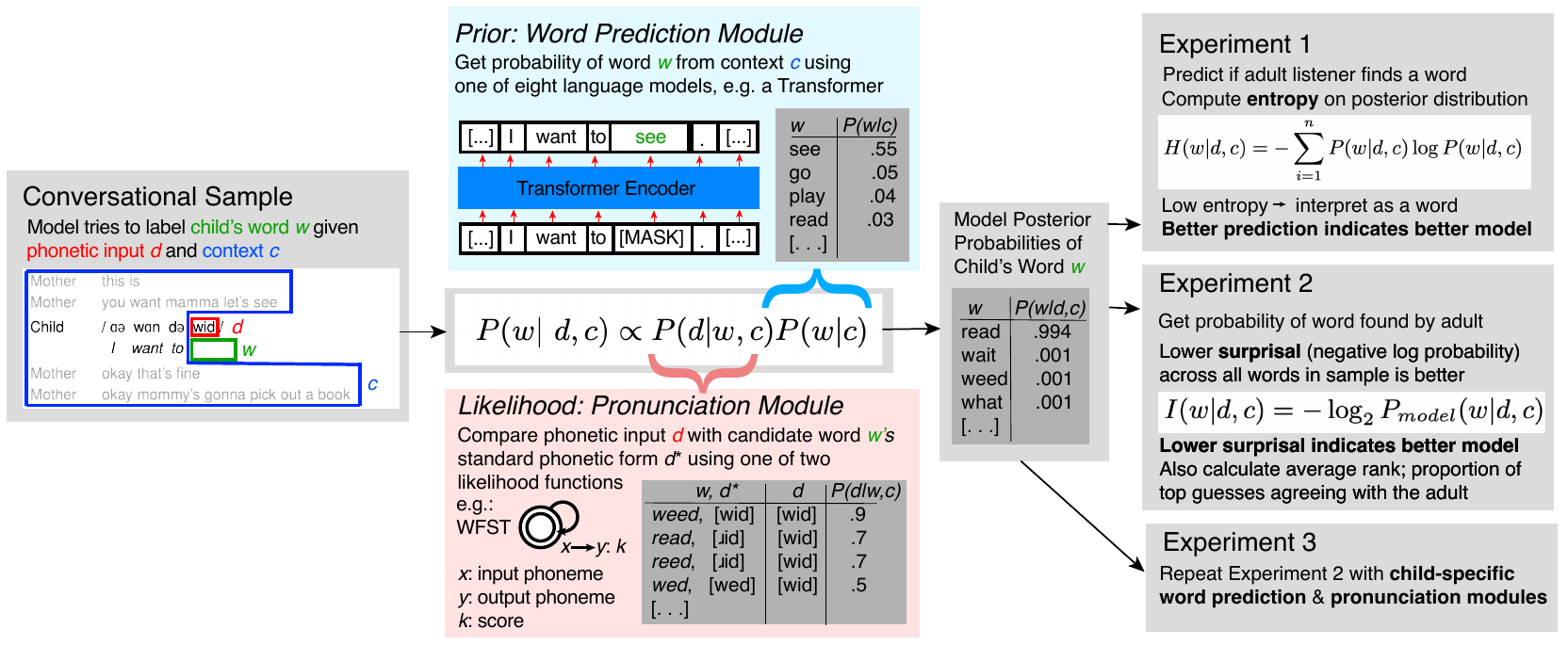}
\caption{
Schematic overview of the Bayesian spoken word recognition models and experiments.
Models take phonetic transcripts of child speech as input and try to predict the adult interpretation using the surrounding context and the phonetic signal provided by the child. 
Each model makes use of (i) a prior (light blue box), in the form of a probabilistic language model that predicts the word identity $w$ based on the context $c$; and (ii) a probabilistic pronunciation model (light pink box). The prior and likelihood are combined using Bayesian inference to form a posterior distribution (light gray box) over candidate words $w$ given $c$ and $d$.
This posterior is evaluated against human annotations in two ways: first, how well it predicts whether an adult listener takes the input to be an intelligible word or treats it as unintelligible (Expt. 1); second, for intelligible input, how well it predicts which word is found by the adult listener (Expt. 2).
We then test models tuned to specific children to see if adults tune to aspects of the speech of specific children (Expt. 3).
}
\label{fig:overview_figure}
\vspace{-5mm}
\end{figure*}

The proposed child-directed listening hypothesis is closely related to ``noisy-channel'' theories of communication \citep{shannon1951} where listeners overcome the imperfect acoustic information, linguistic ambiguity, and speaker variability present in everyday conversation by relying on their constantly-updating expectations about what speakers are likely to say and by taking into account the nature of noise \citep{levy2008noisy, gibsonBergenPiantadosi2013, MeylanNairGriffiths2021}.\footnote{The proposal of child-directed listening differs from standard theories of noisy channel communication in that it does \textit{not} assume that the speaker (the child) is a noisy-channel speaker. 
Instead, the child may not have a message in the adult language in mind when producing a vocalization but the adult listener may interpret that vocalization as a word regardless.}
Specifically, adult listeners consider possible interpretations of the phonetic input they receive in terms of both the interpretation's plausibility in context and its perceptual similarity to typical adult pronunciations.
We thus adopt a Bayesian computational approach to spoken word recognition \citep{levy2008noisy, norrisMcQueen2008, MeylanNairGriffiths2021} to create models that ``guess'' what the child said---whether or not it is a word (Expt 1, see Fig. \ref{fig:overview_figure}), and if it is, \textit{what} word (Expt 2).
We then test for evidence of adults adapting to the speech of specific children (Expt 3).

All proposed models evaluate evidence for different hypotheses about what word a child said (each possible word interpretation, denoted $w$; see Fig.~\ref{fig:overview_figure}) in a particular context (denoted $c$), given the often-imperfect sequence of sounds they produced, which we represent as a string of phonetic segments (\eg [k\ae] for a production of \textit{cat} in which the final /t/ is not pronounced; denoted as data, $d$).
The probability of each interpretation given what the child said is computed with Bayes' rule \citep{chater2008probabilistic, perfors2011tutorial}:
$p(w|d,c) \propto p(d|w,c)p(w|c)$.
This comprises two parts:
each interpretation's \textit{prior} probability, $p(w|c)$---or how likely the interpretation is \textit{before} hearing the child production---and each interpretation's \textit{likelihood}, $p(d|w,c)$---or the fit of the acoustics of the child's production to the word interpretation under consideration. 
We estimate the prior $p(w|c)$ with several different probabilistic language models that predicts the identity of a missing word from the linguistic context---to distinguish this from the overall Bayesian model, we call this a ``word prediction module.'' 
The likelihood, $p(d|w,c)$, can be realized as a ``pronunciation module,'' which captures the probability distribution over possible pronunciations from the child.
Each model considers the 8,000 most common words in the child language environment as possible interpretations (see Methods).

Empirical work has revealed that adults make use of a broad variety of information to infer word identities in spoken language from other adults: among them, the relative frequency of words and phrases \citep{millerEtAl1951, howes1957}, rich representations of discourse and linguistic structure \citep{norrisMcQueenCutler2016, norrisMcQueen2008, rohdeEttlinger2012}, knowledge of available referents in the scene \citep{altmannKamide1999, kamide2003time, tanenhaus1995}, and rapid adaptation to the pronunciations of specific speakers \citep{KleinschmidtJaeger2015}. 
Our \textbf{full model} of adults' understanding of children's speech incorporates all of these information sources.
This involves modeling children's likely (mis)pronounciations, and expectations for the sorts of words, multi-word phrases, and topics that children are likely to talk about. 
To analyze the relative contributions of the components of this full model,
we adopt the machine learning technique of \textit{model ablation}: we test the effects of systematically substituting simpler baselines for different components of the model to assess the effects on performance \citep{reddy1975speech}.
This gives evidence about what information sources are most important for reproducing adult-like listening behaviors. 

\textbf{\textit{Handling Child Mispronunciations}} The full model learns how children are likely to mispronounce words, \eg that some mispronunciations are more likely for children than others, by estimating phoneme-specific deletion, insertion, and substitution probabilities using a weighted finite state transducer (Phoneme-Specific likelihood) in its pronunciation module (see Methods). 
We compare this approach with an ablated variant in which all deletions, insertions, and substitutions of phonemes are assumed to be equally likely (Edit-Distance likelihood, following \citep{wagner1974string}). 
This reveals the degree to which adults use regularities in children's pronunciations to understand their speech.

\textbf{\textit{Word Prediction Architecture}} We test a wide range of model variants for the word prediction module, following the naming scheme \texttt{\{model architecture\} + \{training dataset\} + \{use of context\}} (note that the latter two dimensions are not applicable for some architectures, in which case the labels are omitted).
The full model uses \textsc{BERT}, an advanced neural network architecture that is able to represent rich semantic and syntactic relationships between words and can predict the identity of missing (or ``masked'') words \citep{devlin2018}.
\textsc{BERT} uses a ``bidirectional'' input, or words from either side of the target word (here we use up to 256 words in either direction, which easily accommodates 20 utterances). 
In addition to this representationally rich, bidirectional model, we investigate four ablations in the model architecture component of the word prediction module, allowing us to parse apart the contributions of tracking alternating dyadic turn structure in conversation, topics and anaphora, long-distance syntactic dependencies, word sequences, and (at the most basic level) word frequencies.
First, we test an alternative neural network architecture, \textsc{GPT-2} \citep{radford2019language}, which uses only preceding words for prediction and may be a yet better model of realtime human language prediction  (\citep{meister2021revisiting, schrimpf2021neural}; see Use of Context in Word Prediction, below).
Second, we evaluate the predictive utility of a simpler baseline trigram model \citep{manning1999foundations} fit on the same corpus of child speech in two ways. 
Unlike the transformer models above, the trigram models are not able to track long-distance dependencies between words.
We use a \textsc{Trigram} model to estimate the probability of each word in the probability as a continuation given the two previous words (the relevant dataset used to fit the trigram model is described below).
Third, we evaluate a model that reflects only the relative frequency of use of each word in the a large corpus of child speech (\textsc{Unigram}).
Fourth, we evaluate the performance of a model that assumes \textit{a priori} that all words in the vocabulary are equally likely (\textsc{UniformPrior}).
This final model represents a critical baseline in that it is entirely dependent on the child's pronunciation to determine the child's intended word: all other models reflect prior knowledge about what children are likely to say rather than relying directly on the acoustic signal supplied by the child.
Testing these ablated versions clarifies the relative importance and nature of adult expectations regarding language structure when interpreting children's speech. 

\textbf{\textit{Fine-Tuning Word Prediction}} 
Where possible, we fine-tune or fit word prediction models using a large dataset of transcripts of child-caregiver interactions in the home environment in the same age range as the test set, taken from the CHILDES corpus
\citep{macwhinney2000childes}.
These models are designated by \textsc{+CHILDES}.
Fine-tuning and model fitting procedures vary by model and are detailed in the Methods.
Among the BERT models, we test the consequences of changing the data used to fine-tune the model in two ablations.
In the first, we fine-tune the model on a corpus of adult-to-adult telephone conversations (\textsc{+Switchboard}; \citep{godfrey1992switchboard}).
In the second, we do not fine-tune the model at all, leaving it in its base form where it was trained on a large collection of Internet-based written texts, particularly news articles articles and web forums (\textsc{+AdultWritten}).
This comparison provides insight into how expectations regarding how young children speak may differ from other kinds of language.

\textbf{\textit{Use of Context in Word Prediction}} The full model takes into account the content of up to 20 preceding and following utterances when predicting the identity of the missing word (\textsc{+Bidirectional}).
Because we evaluate these models against word identities found by adult annotators who can consult video and audio from utterances following the target word, we expect the best performance from a model that can use both preceding and following utterances to predict the missing word. 
Looking at both preceding and following utterances means that if a potential referent (\eg a person, animal, object, action, etc.) is mentioned in the following context, the model places a higher probability on it as the identity of the target word.
This differs from the situation faced by caregivers, who---unlike annotators---must identify the word largely on the basis of the content of preceding utterances.
For this reason we investigate a set of ablated model variants that use the 20 preceding utterances for prediction (\textsc{+Preceding}).\footnote{BERT models are trained and fine-tuned using a bidirectional masked word prediction task and the availability of context is only restricted at inference time. We use the bidirectional task for training these models for the pragmatic reason that this is known to yield very useful linguistic representations, and we make no claims regarding the appropriateness of BERT as a learning model for the adult. GPT-2-based models, by contrast, are trained and fine-tuned with a left-to-right prediction task.} 
BERT's bidirectional architecture means that in the \textsc{+Preceding} ablation, preceding utterances as well as both preceding and following words in the utterance with the target word are used for prediction; for GPT-2, prediction is based on preceding words only.
We also test a set of ablated model variants that use only the immediate utterance for word prediction (\textsc{+OneUtt}).
This set of model ablations clarifies to what extent adults depend on the larger linguistic context in interpreting children's speech.





In broad strokes (detailed below), our analyses reveal that the full model (\textsc{BERT+CHILDES+Bidirectional}) achieves a strong quantitative fit to adult interpretations of children's speech, accurately predicting the word identity in more than 90\% of cases, even when considering a vocabulary of approximately 8,000 possible word identities.
This best-performing full model has strong priors --- expectations that children will be generally more likely to: 
produce grammatical utterances (\eg \textit{two dogs} rather than \textit{two dog}), refer to people and things in the broader discourse context  (\eg \textit{cat} if a cat has been recently mentioned), and talk about child-relevant words (\eg \textit{read} rather than \textit{weed}).
Model ablations reveal that adapting to child pronunciations alone or failing to use contextual cues to identify words yields a poor fit to the words found by adults.  
These analyses also provide evidence that adults tune their listening to the speech of individual children. 
We conclude by discussing how adults' inferences that go beyond children's phonetic productions may provide critical support for early dyadic caregiver-child communication and how this in turn may drive the incremental refinement of children's earliest language knowledge. 

\section*{Results}

\subsection*{Expt. 1: Predicting Whether Adults Find Words} 
We first tested which spoken word recognition model best predicts when adults find word interpretations in children's speech vs. when they consider that speech to be unintelligible.
We did this by evaluating each model's ability to predict when adults flagged word-length segments of children's speech as unintelligible, vs. identify them as specific words in a large corpus of child-produced speech. 
As a continuous predictor of intelligibility, we used each models' estimates of entropy over word identity (Fig. \ref{fig:overview_figure}, Expt. 1) and used this as the sole input to a binary classifier.

\begin{figure*}
    \centering
    \begin{overpic}[width=0.43\textwidth]{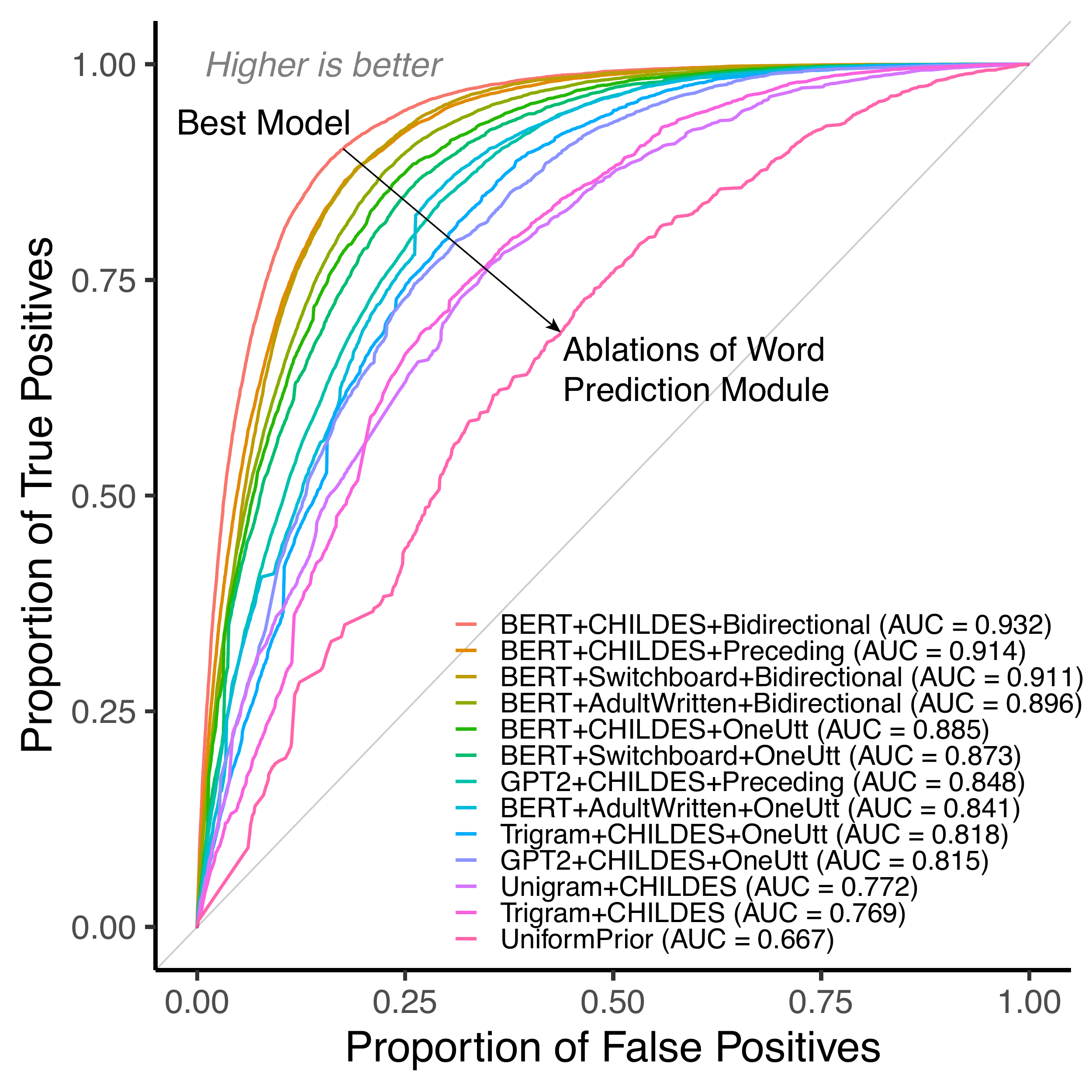}
 \put (-5,95) {\Large \textsf{A}}
    \end{overpic}
    \hspace{2em}
    \begin{overpic}[width=0.43\textwidth]{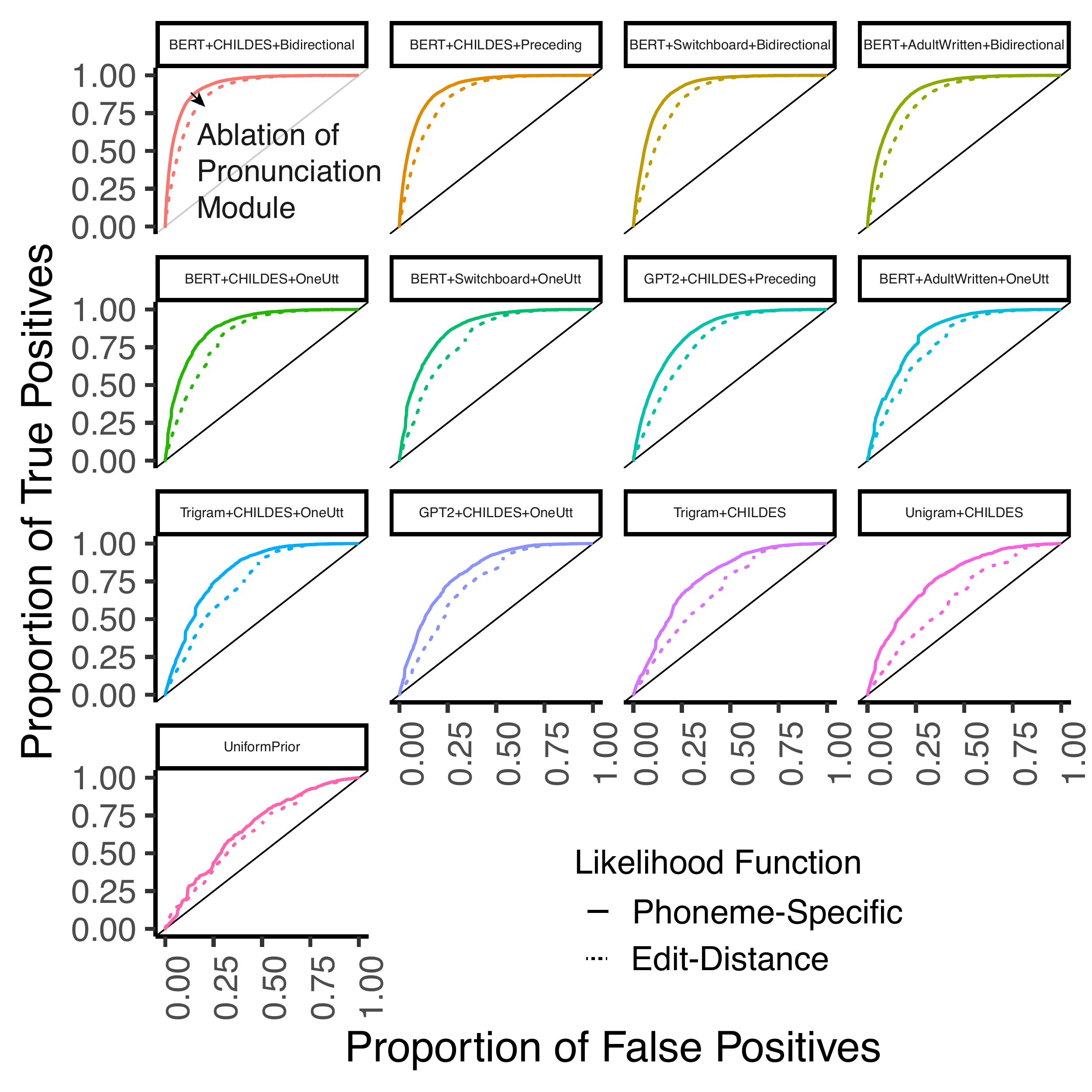}
 \put (-5,95) {\Large \textsf{B}}
    \end{overpic}
    
    \caption{\textbf{A.} Classification performance by model when predicting if adults will provide a word interpretation versus flag a vocalization as unintelligible ($n = 57,812$ and $n=12,786$, respectively).
    In panel A, all models use the Phoneme-Specific likelihood in the pronunciation module. 
    The solid line with slope = 1 indicates chance performance.
    A larger area under the curve (AUC) corresponds to better classification performance. 
    \textbf{B.} 
    Classification performance on the same task comparing models using the Phoneme-Specific likelihood (solid lines) with those using the ablated Edit-Distance likelihood (dotted lines).
    Priors are matched within each panel.
    Every model that uses a Phoneme-Specific likelihood outperforms the analogous one using the Edit-Distance likelihood.
    }
    \label{fig:roc_comparison}
    
\end{figure*} 

\subsubsection*{Prior: Comparing word prediction modules}
We evaluated the statistical significance of differences in classifier performance, operationalized as area-under-the-curve (AUC), and assessed the statistical significance in pairwise comaparison of model performance using DeLong's test \citep{delong1988comparing}.
The full model (\textsc{BERT+CHILDES+Bidirectional}), which reflects rich linguistic beliefs tuned specifically to child language productions, had the highest AUC (.932); this model best replicates the behavior of adult listeners in predicting whether adults found a given vocalization to be intelligible (Figure~\ref{fig:roc_comparison}A).
This higher performance is statistically significant in comparison to all other models.
In fact, most pairwise comparisons between all models were significant, with the exception of \textsc{Trigram+CHILDES} vs. \textsc{Unigram+CHILDES} ($Z = 0.772, p = 0.44$) and \textsc{GPT2+CHILDES+OneUtt} vs. \textsc{Trigram+CHILDES+OneUtt} ($Z = 1.773, p = 0.076$).
The \textsc{UniformPrior} performs substantially worse than all others.
Under this model, prior entropy is constant across all word types, consequently this classifier relies heavily on the contribution of the phonetic data produced by the child, and it performs much closer to chance in predicting whether words were found to be intelligible.
This analysis provides evidence that a prior that is tuned specifically to child language and uses the surrounding utterance context (\ie our \textsc{BERT+CHILDES+Bidirectional} model) is best able to replicate adult inferences, particularly as to which vocalizations are likely to be interpreted as words by adults.

\subsubsection*{Likelihood: Comparing pronunciation modules}

Across all word prediction model architectures, the Phoneme-Specific likelihood outperforms the Levenshtein distance-based likelihood in predicting adult listening behavior (Fig.~ \ref{fig:roc_comparison}B).
However, the performance difference between pronunciation modules for a given word prediction module is much less pronounced than the performance difference across word prediction modules for a given pronunciation module (compare Fig.~ \ref{fig:roc_comparison}B vs. Fig.~ \ref{fig:roc_comparison}A).
We investigate the reasons and quantitative difference in greater detail in Expt. 2.

\subsection*{Expt. 2: Predicting the Words Recovered By Adults} 
While Expt. 1 compares model performance in determining \textit{whether} adults would find a child production to be interpretable as a word or not, Expt. 2 digs into the specifics of word identity, asking whether our Bayesian spoken word recognition models recover the same words as adult transcribers. 
This reflects the idea that models must not only effectively predict what will be treated as a word, but also how adults will interpret a given vocalization.
To evaluate the degree to which models replicate adult listening behavior, we focus on the cases where transcribers interpreted children's productions as transcribable words and compute the surprisal \citep{levy2008Expectation, hale2001probabilistic} or self-information that each model assigns to the transcriber's interpretation (see Methods).
The lower the average surprisal of the transcriber's word interpretation, the better the model reflects adult inferences about child speech.
In addition, we report the average rank of the transcriber's word interpretations for each model (ideal model performance would result in a mean rank of 1, which would mean that the transcriber's word is always the highest ranked word interpretation according to the model).
We also report the percentage of vocalizations in the test set where the model's highest probability guess was the same word found by the adult (``Correct Top Guess'', Table \ref{tab:model_surprisal_comparison}).
Here we report model performance across a test set taken from all children in the Providence corpus; for a follow-up analysis of model performance by child, see \textit{Supplementary Note: Spoken Word Recognition Model Performance By Child}.

\newpage
\begin{table*}[h]
\tablefont
\caption{Model performance as measured by average surprisal, average rank of transcriber interpretation, and percentage of correct top guesses for word recognition models.
The  left column (``Prior'') reflects the distributions from the word prediction module, \ie before using the child's production to identify the word.
The center and right columns reflect integrating information from two variants of the pronunciation module.
The full model is indicated in red, which we expect to best replicate adult interpretations of child speech. 
We compare this with a phonetics-only baseline model, which must recognize words solely on the basis of the acoustic signal produced by the child (green).
Ablated models test which information sources are most critical for the full model.
\vspace{-5mm}
}
\label{tab:model_surprisal_comparison}
\begin{center}
\begin{tabular}{l|lll||lll|lll}
&  \multicolumn{3}{c||}{Prior} & \multicolumn{6}{c}{Posterior} \\

& \multicolumn{3}{c||}{} & \multicolumn{3}{c|}{Phoneme-Specific Likelihood} & \multicolumn{3}{c}{Edit-Distance Likelihood} \\
\cline{2-10}
& Average     & Average       & Correct       & Average    & Average & Correct & Average  & Average & Correct  \\
& Surprisal$^\star$     & Rank$^\ddagger$ & Top  & Surprisal$^\star$ & Rank$^\ddagger$ &
Top & Surprisal$^\star$       & Rank$^\ddagger$ & Top       \\
Language Model Prior             & (bits$^\dagger$)        &    &  Guess$^\Diamond$  & (bits$^\dagger$) &         &  Guess$^\Diamond$ & (bits$^\dagger$)          &      &  Guess$^\Diamond$ \\ \hline
& & & & \multicolumn{3}{c|}{\cellcolor{red!25} Full Model} & \multicolumn{3}{c}{\cellcolor{blue!25} Ablated Models} \\

\textsc{BERT+CHILDES+Bidirectional} & 2.78 & 30.76 & 60\% & \cellcolor{red!25} \textbf{0.57} & \cellcolor{red!25} \textbf{1.80} & \cellcolor{red!25} \textbf{90\%} & \cellcolor{blue!25} 1.03 & \cellcolor{blue!25} 5.63 & \cellcolor{blue!25} 84\%  \\

\textsc{BERT+Switchboard+Bidirectional} & 3.41          & 49.77   & 54\%   & \cellcolor{blue!25} 0.74            & \cellcolor{blue!25} 3.12 & \cellcolor{blue!25} 88\% &  \cellcolor{blue!25} 1.32  & \cellcolor{blue!25} 9.19  & \cellcolor{blue!25} 81\%     \\

\textsc{BERT+CHILDES+Preceding}     & 3.56  & 61.39   & 52\%  & \cellcolor{blue!25} 0.77 & \cellcolor{blue!25} 3.52  & \cellcolor{blue!25} 88\%   & \cellcolor{blue!25} 1.41   & \cellcolor{blue!25} 11.85 & \cellcolor{blue!25} 80\%  \\

\textsc{BERT+AdultWritten+Bidirectional}     & 4.76          & 91.47    & 39\%  & \cellcolor{blue!25} 1.11            & \cellcolor{blue!25} 6.05  & \cellcolor{blue!25} 83\%   & \cellcolor{blue!25} 1.85   & \cellcolor{blue!25} 18.25 & \cellcolor{blue!25} 73\%  \\

\textsc{BERT+CHILDES+OneUtt}  & 5.02          & 124.52  & 36\%   & \cellcolor{blue!25} 1.18            & \cellcolor{blue!25} 7.61 & \cellcolor{blue!25} 82\%  & \cellcolor{blue!25} 2.05 & \cellcolor{blue!25} 24.35 & \cellcolor{blue!25} 72\%\\

\textsc{BERT+Switchboard+OneUtt}  & 6.33          & 342.27  & 28\%   & \cellcolor{blue!25} 1.60            & \cellcolor{blue!25} 15.84 & \cellcolor{blue!25} 78\% & \cellcolor{blue!25} 2.63 & \cellcolor{blue!25} 44.11 & \cellcolor{blue!25} 67\%\\

\textsc{GPT2+CHILDES+Preceding}     & 6.45  & 396.27    & 29\%  & \cellcolor{blue!25} 2.01 & \cellcolor{blue!25} 25.91  & \cellcolor{blue!25} 75\%   & \cellcolor{blue!25} 3.17   & \cellcolor{blue!25} 66.76 & \cellcolor{blue!25} 63\%  \\

\textsc{BERT+AdultWritten+OneUtt}      & 7.16          & 459.78    & 26\% & \cellcolor{blue!25} 1.90            & \cellcolor{blue!25} 18.74  & \cellcolor{blue!25} 74\%   &  \cellcolor{blue!25} 2.92 & \cellcolor{blue!25} 48.8 & \cellcolor{blue!25} 65\%\\

\textsc{Trigram+CHILDES} & 7.82  & 679.72  & 14\%  & \cellcolor{blue!25} 2.54 & \cellcolor{blue!25} 21.21  & \cellcolor{blue!25} 65\%   & \cellcolor{blue!25} 3.73   & \cellcolor{blue!25} 57.3 & \cellcolor{blue!25} 54\%  \\

\textsc{GPT2+CHILDES+OneUtt}     & 7.88  & 549.98    & 18\%  & \cellcolor{blue!25} 2.60  & \cellcolor{blue!25} 45.41  & \cellcolor{blue!25} 71\%   & \cellcolor{blue!25} 3.99   & \cellcolor{blue!25} 104.85 & \cellcolor{blue!25} 58\%  \\

\textsc{Trigram+CHILDES+OneUtt}  & 8.04  & 490.70  & 29\%  & \cellcolor{blue!25} 1.93 & \cellcolor{blue!25} 15.29  & \cellcolor{blue!25} 73\%   & \cellcolor{blue!25} 2.89   & \cellcolor{blue!25} 43.39 & \cellcolor{blue!25} 63\%  \\

\textsc{Unigram+CHILDES}   & 8.74          & 305.69    & 3\%  & \cellcolor{blue!25} 2.36 & \cellcolor{blue!25} 16.86  & \cellcolor{blue!25} 66\%   &  \cellcolor{blue!25} 3.91 & \cellcolor{blue!25} 50.79 & \cellcolor{blue!25} 49\% \\

\textsc{UniformPrior}    & 12.97         & 3998.00 & 0\%      & \cellcolor{green!25} 4.13           & \cellcolor{green!25} 49.1 & \cellcolor{green!25} 42\% & \cellcolor{blue!25} 5.16 & \cellcolor{blue!25} 143.26 & \cellcolor{blue!25} 30\% \\

& & &  & \multicolumn{3}{c}{\cellcolor{green!25} Phonetics-Only Baseline Model} & &  & \\



\end{tabular}
\end{center}
\vspace{-3mm}
\footnotesize{\noindent $^\star$ A lower surprisal indicates a better model. The difference in average probability assigned to the transcriber interpretation is $2^{\text{diff}}$, where $\text{diff}$ is the difference between two model scores. \\
$^\dagger$ Paired \textit{t}-tests yield statistically significant pairwise differences between all model surprisal values, $p<10^{-5}$. \\
$^\ddagger$ A lower average rank indicates a better model. Ranks are taken out of a vocabulary of 7,997 possible words.\\
$^\Diamond$ A higher percentage of correct top guess indicates a better model.
}
\end{table*}


\subsubsection*{Prior: Comparing word prediction modules}
The rows of Table~\ref{tab:model_surprisal_comparison} show comparisons of models using different priors (paired \textit{t}-tests yield statistically significant  differences between all adjacently ranked model pairs, all $p<0.0001$ after Bonferroni correction for multiple comparisons).
Among the ablated models, those using the preceding and following linguistic context (\textsc{+Bidirectional}), including the full model, systematically outperform the ones that use only the immediate utterance for context (\textsc{+OneUtt}). 
Using only the preceding and current utterance for prediction (\textsc{BERT+CHILDES+Preceding}) only minimally impacts model performance relative to the bidirectional context, suggesting that the model can perform well using information available to caregivers in realtime.
Word prediction modules that use the \textsc{BERT} architecture generally outperform  \textsc{GPT-2}-based models, which in turn outperform the \textsc{Trigram} models,\footnote{\textsc{Trigram+CHILDES+OneUtt} shows an outlying pattern of relatively high average surprisal (Table 1), yet a much more competitive average rank, percentage correct top guess, and yields highly competitive posteriors. Follow-up analysis reveals a much flatter probability distribution over candidate interpretations than the other models, though the relative rankings of candidates are a reasonable match to annotators.} which in turn outperform \textsc{Unigram}.
While \textsc{BERT+Preceding} and \textsc{GPT-2+Preceding} models both use Transformer-based architectures and largely or entirely rely on the preceding context, the former show uniformly higher performance, possibly because of the additional information available from same-utterance context immediately following the target vocalization.
All models that fit prior expectations to data in some way substantively outperform the baseline \textsc{UniformPrior} model.
Additionally, BERT models fine-tuned on child speech (\textsc{+CHILDES}) outperform ablated models fine-tuned on adult speech (\textsc{+Switchboard}), which in turn outperform models based on written text (\textsc{+AdultWritten}). More broadly, we see that the more context available to the model, the more sophisticated the model architecture (neural network vs.\ trigram), and the more tuned the model to the type of language in our dataset (spoken language produced by children), the better the model performs in predicting how adults interpret child vocalizations.\footnote{One exception to this generalization is the notably low performance of \textsc{GPT2+CHILDES+OneUtt}, which is below some or all \textsc{Trigram} models depending on the evaluation metric used. One possible explanation is that because most child utterances are short, a considerable proportion of target vocalizations are very near the beginning of an utterance. When the \textsc{GPT2} architecture is constrained to same-utterance left context, there is little information to predict from.}

\begin{figure}[t]
\centering
\includegraphics[width=.75\linewidth]{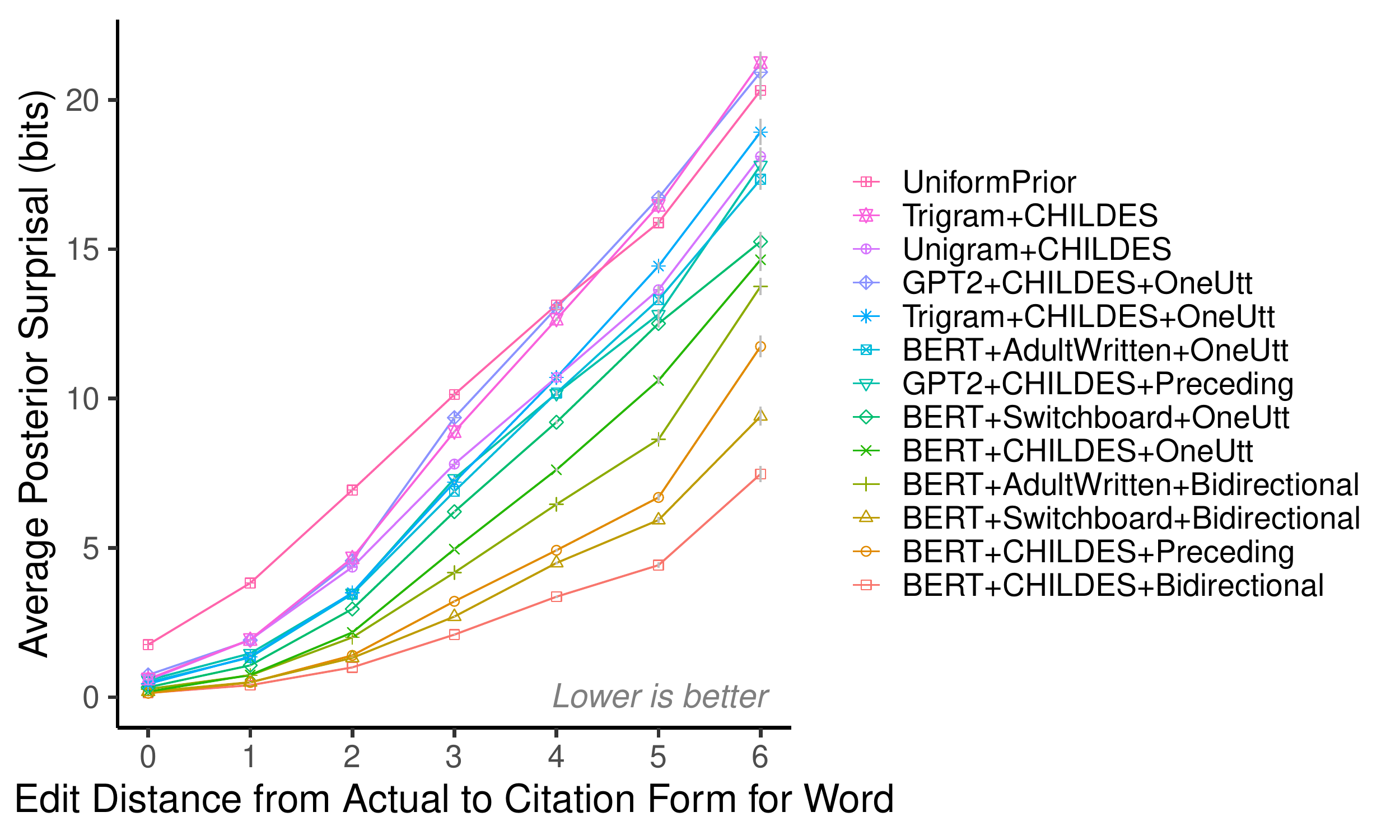}
\caption{
Average posterior surprisal of the transcribers' recovered word interpretation under each model with the phoneme-based likelihood, plotted  according to the similarity of the child production to the standard adult pronunciation (x-axis). 
Vertical error bars (gray) indicate the standard error of the mean; at many edit distances the SEM is too small to be seen.
All models yield comparable predictions for vocalizations that are similar to adult pronunciations (left side of the graph). 
However, models with sophisticated knowledge of syntax and discourse structure, which are fine-tuned to child language and consider the broader linguistic context, assign a much lower surprisal (higher probability) to the words recovered by transcribers when pronunciations deviate from the adult language (right of the graph).
This points to how the contribution of adults' prior expectations increases with the degree of child mispronunciation.
}
\label{fig:posterior_entropy_by_edit_distance}
\end{figure}

\subsubsection*{Likelihood: Comparing pronunciation modules}
Integrating information from children's pronunciations through the two likelihoods results in a better fit to the adult interpretations of children's speech, but maintains the same relative ranking of performance as seen from the word prediction modules (Table~\ref{tab:model_surprisal_comparison}).
Indeed, the contribution of the model priors / word prediction module to the model posteriors is so critical that the full model (\textsc{BERT+CHILDES+Bidirectional}) word prediction module makes \textit{better} inferences about the word interpretation \textit{without} using the phoneme string produced by the child (2.78 bits surprisal, average rank 30.76, 60\% correct top guess) than a model with minimally informative prior word expectations that does use the phonetic form (\textsc{UniformPrior} with the Phoneme-Specific likelihood; 4.13 bits surprisal, average rank 49.1, 42\% correct top guess).
The full model (which uses the \textsc{BERT+CHILDES+Bidirectional} word prediction module, plus the Phoneme-Specific likelihood) outperforms the ablated model that uses the Levenshtein distance likelihood (0.57 bits of surprisal, average rank of 1.80, 90\% correct top guess vs.  1.03 bits of surprisal, average rank of 5.63, 84\% correct top guess).
The full model also outperforms the ablated model where the word prediction module is fine-tuned on adult-to-adult conversational speech (\textsc{Switchboard} with the Phoneme-Specific likelihood, 0.74 bits of surprisal, average rank of 3.12, 88\% correct top guess).
All these models substantively outperform ones with less sophisticated expectations about language structure (\eg \textsc{Unigram+CHILDES} with the Phoneme-Specific prounciation module, average surprisal of 2.36 bits, average rank of 16.86, and 66\% correct top guess).
The much lower performance of the model with the\textsc{UniformPrior} and
Phoneme-Specific pronunciation module (42\% correct top guess) reveals that children's pronunciations alone are insufficient for inferring their intended word identities, and makes clear the importance of prior expectations for replicating adult interpretations of children's speech.

\begin{figure*}
    \centering
    \begin{overpic}[width=0.43\textwidth]{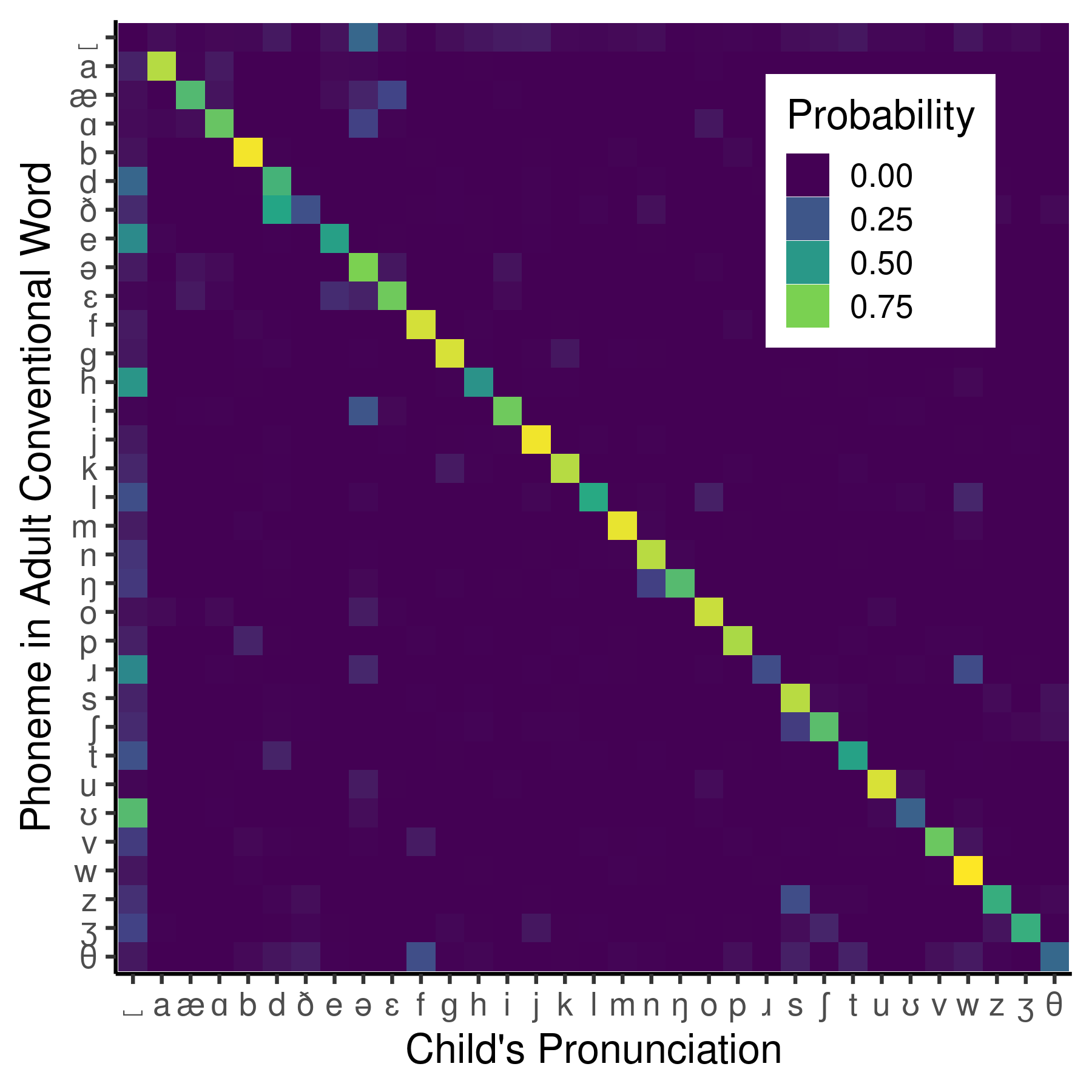}
 \put (0,95) {\Large \textsf{A}}
    \end{overpic}
    \hspace{5em}
    \begin{overpic}[width=0.43\textwidth]{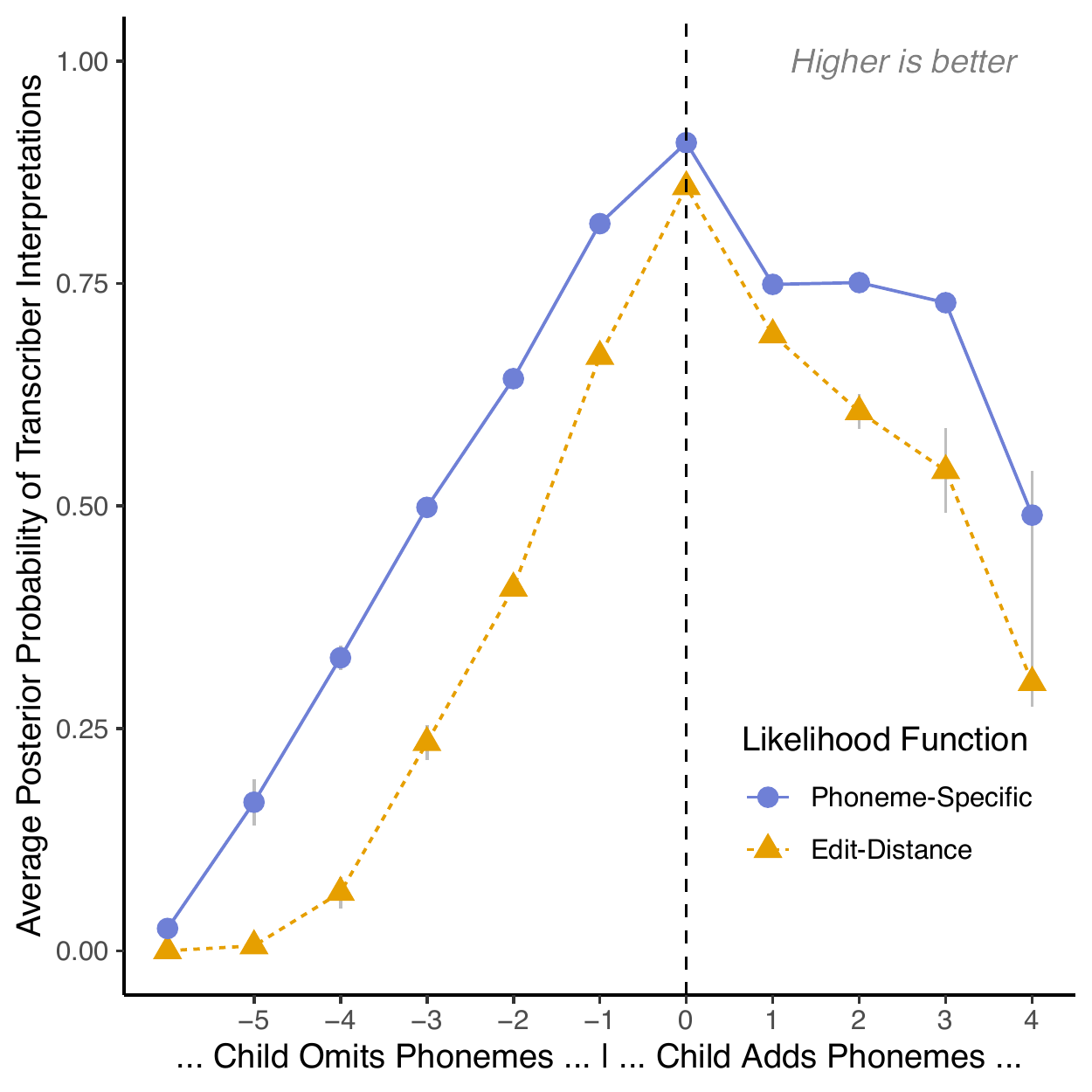}
 \put (15,95) {\Large \textsf{B}}
    \end{overpic}
    
    \caption{\textbf{A.} Substitution probabilities for the Phoneme-Specific likelihood. \texttt{\verbvisiblespace}$ \rightarrow \{\text{phoneme}\}$  indicates insertion (top row), while $\{\text{phoneme}\} \rightarrow$ \texttt{\verbvisiblespace} indicates deletion (leftmost column). 
    Off-diagonal entries reflect common deviations from adult pronunciations made by English-learning children: 
    [f] in place of [\textipa{T}] (\textit{think} $\rightarrow$ \textit{fink}), 
    [w] in place of [\textipa{\*r}] (\textit{read} $\rightarrow$ \textit{weed}),
    [s] in place of [\textipa{S}] (\textit{fish} $\rightarrow$ \textit{fiss}).
    The leftmost column captures that children are likely to delete many phonemes altogether.
    \textbf{B.} Average posterior probability of words found by adult transcribers under two variants of the \textsc{BERT+CHILDES+Bidirectional} model: one using the Phoneme-Specific likelihood and one using the Edit-Distance likelihood (error bars, where visible, reflect the standard error of the mean).
    While the two variants show comparable performance in cases where children include unexpected additional sounds in their pronunciation, the one using the Phoneme-Specific likelihood is a better fit to adult data when children drop many of the phonemes that make up the conventional pronunciation of the word (left side of graph).
    }
    \label{fig:fst_analysis}
\end{figure*}    

To better understand the contribution of priors expectations, we investigated their contribution as a function of how non-standard the child's pronunciation was in comparison to the standard adult form for the word found by transcribers.
We then compared a linear regression predicting posterior surprisal using 
the interaction of model and edit distance versus one using only edit distance.
The former model substantially outperforms the latter (BIC = 5,200,682 vs. BIC = 7,673,790), suggesting that models differ in their ability to replicate adult interpretations depending on the fidelity of the child pronunciation to the adult form.
While most models assign high probabilities to the word found by adults when the child's pronunciation is close to the adult form, the BERT-based priors are a much better match to adults when the child's production is relatively \textit{dissimilar}  (right side of Figure~\ref{fig:posterior_entropy_by_edit_distance}).
For example, for child productions that are two edits away from the adult form ($x = 2$ in Fig. \ref{fig:posterior_entropy_by_edit_distance}), the full model (\textsc{BERT+CHILDES+Bidirectional}) assigns approximately 64 times the probability to the transcribers' interpretations compared to the model that only uses the phonetic input \textsc{UniformPrior}.
This difference in probabilities becomes yet more pronounced at higher edit distances: at six edits, the full model assigns transcribers' interpretations 5,928 times the probability as the \textsc{UniformPrior} model.
This analysis provides clear evidence that adult listeners rely most heavily on their prior expectations when interpreting child vocalizations that differ the most from words in the adult language. 

While the above analyses highlight the critical importance of the prior regardless of the choice of likelihood, we sought to further understand why models using the the Phoneme-Specific likelihood, including the full model, consistently outperform those using the Edit-Distance one.
We visualize the probability over phoneme-level edits from the fitted WFST from the Phoneme-Specific likelihood as a transition matrix (Figure~\ref{fig:fst_analysis}A).
Most phonemes are likely to be pronounced correctly by children, as captured by the highest values along the diagonal.
However, English-learning children are very likely to omit many phonemes (leftmost column in Figure~\ref{fig:fst_analysis}A).
This contrasts with the Edit-Distance likelihood, which considers insertions and deletions as equally probable.
Nonzero values off of the diagonal capture many common approximations in the speech of English-learning children:  [f] in place of [\textipa{T}] (\textit{think} $\rightarrow$ \textit{fink}),  [w] in place of [\textipa{\*r}] (\textit{read} $\rightarrow$ \textit{weed}), and [s] in place of [\textipa{S}] (\textit{fish} $\rightarrow$ \textit{fiss}).
The model also captures children's propensity to add an unstressed mid central vowel \textipa{@} (i.e., the default vowel in a relaxed vocal tract) to change the prosodic structure of words to make them easier pronounce, especially to break up consonant clusters (\eg pronouncing \textit{black} as \textit{bulack},  [\textipa{bl\ae k}] $\rightarrow$ [\textipa{b@l\ae k}]).

We next tested whether the improvement observed under the Phoneme-Specific likelihood module is driven by its ability to capture children's articulatory approximations or their deletions.
We compute the average posterior probability of transcribers' interpretations stratify by whether the child's pronunciation is shorter, the same length, or longer than the adult citation form (\ie, standard pronunciation).
If the improvement comes from better accounting for the child's articulatory approximations, then the model with the Phoneme-Specific likelihood will show an improvement in performance when the child's form is of comparable length to the adult citation form.
If, by contrast, the improvement comes from expecting children to \textit{delete} many phonemes, then the model should assign a higher posterior probability to the transcriber's interpretations than the Edit-Distance likelihood when the child produces a much \textit{shorter} form.
Figure \ref{fig:fst_analysis}B provides evidence in favor of the latter hypothesis: the improvement of the Phoneme-Specific likelihood over the Edit-Distance one in predicting adult interpretations is greatest when children omitted many phonemes (towards the left of the graph).  

\subsection*{Expt. 3: Fitting Models to Developmental Age and Specific Children} 

\begin{figure}
\centering
\includegraphics[width=.5\linewidth]{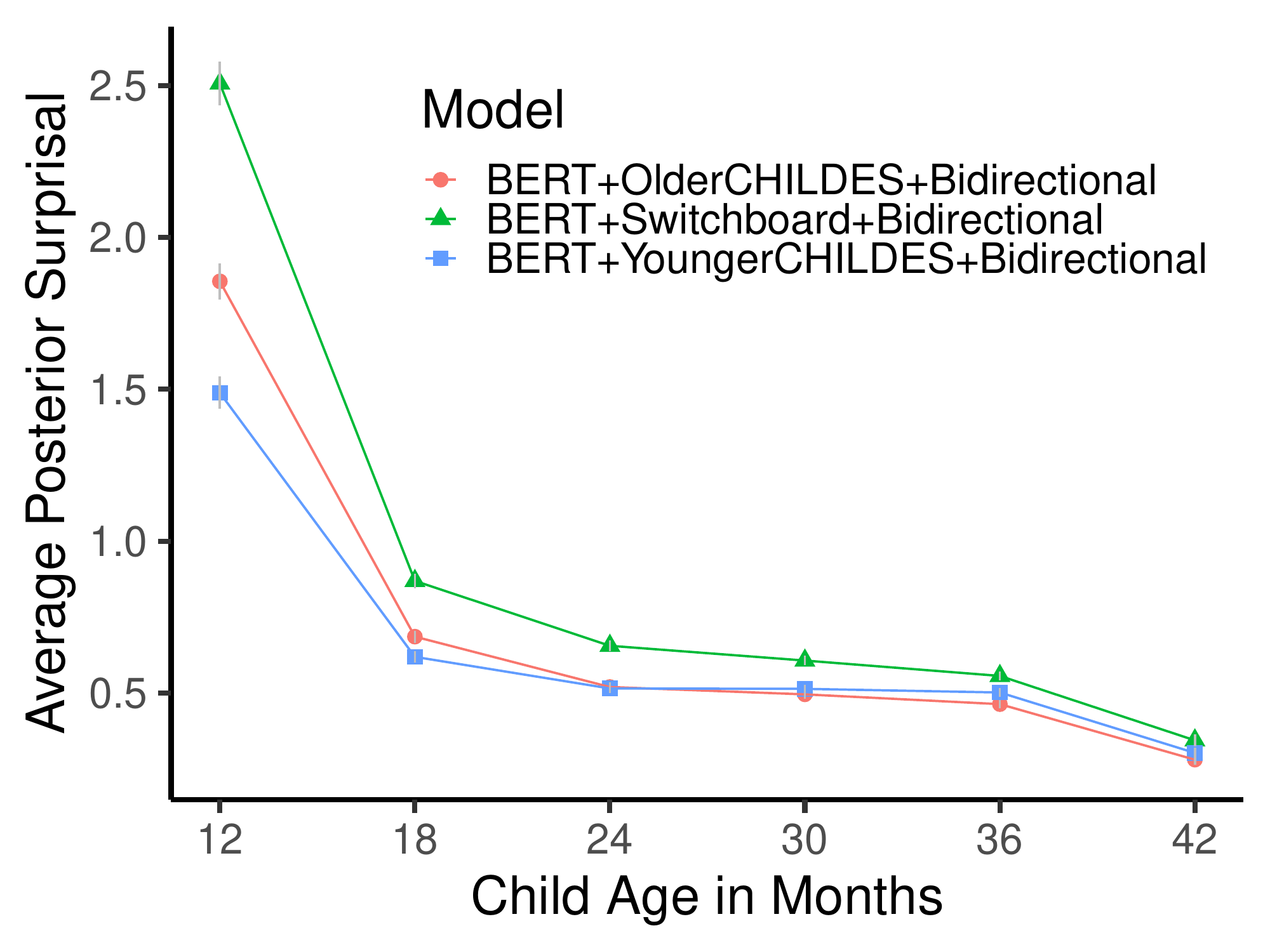}
\caption{
Average posterior surprisal of fine-tuned model guesses by child age.
Relative to a model fine-tuned on conversations with older children (red circles), a model fine-tuned on language usage from younger children (blue squares) does better at recovering transcriber interpretations from young children and slightly worse for those from older ones.
Both outperform a model trained on adult-to-adult conversational speech (green triangles). 
Error bars, where visible, reflect standard errors of the mean; at most ages sample sizes are sufficiently large that standard errors of the mean are too small to be visible on the plot.
}
\label{fig:posterior_surprisal_by_age}
\end{figure}

The results presented so far involve only a small degree of model fine-tuning, to general datasets of child or adult spoken language.
We now further ask whether fine-tuning prior expectations or likelihood functions to specific developmental ages and even to specific children's language production---beyond that proposed in the previously-discussed full model---leads to better predictions regarding what adults will understand.
To test whether additional fine-tuning can capture finer-grained adult expectations, we fine-tune two separate models, one on a language sample for 11-29 month old children and one from 30 to 48 months.
The model trained on the speech of young children (\textsc{BERT+YoungerCHILDES+Bidirectional}) demonstrates better word recognition performance for speech from young children, while a model trained on the speech of older children (\textsc{BERT+OlderCHILDES+Bidirectional}) demonstrates better word recognition on the speech of older children (Fig. \ref{fig:posterior_surprisal_by_age}).
To test the for statistical significance of this difference, we fit a mixed effects model predicting surprisal using the interaction of model type and child age and a maximal random effects structure \citep{barrEtAl2013}.
The age by model type interaction term in this model reveals that the model trained on younger children (\textsc{BERT+YoungerCHILDES+Bidirectional}) assigns higher surprisal to transcriber-found words at older ages ($\beta$ = .184, SE = .013, $Pr(>|t|)$ < $10^{-10}$). 

\begin{figure*}[t]
    \centering
    \begin{overpic}[width=0.48\textwidth, trim=4cm 3cm 5cm 0cm,clip]{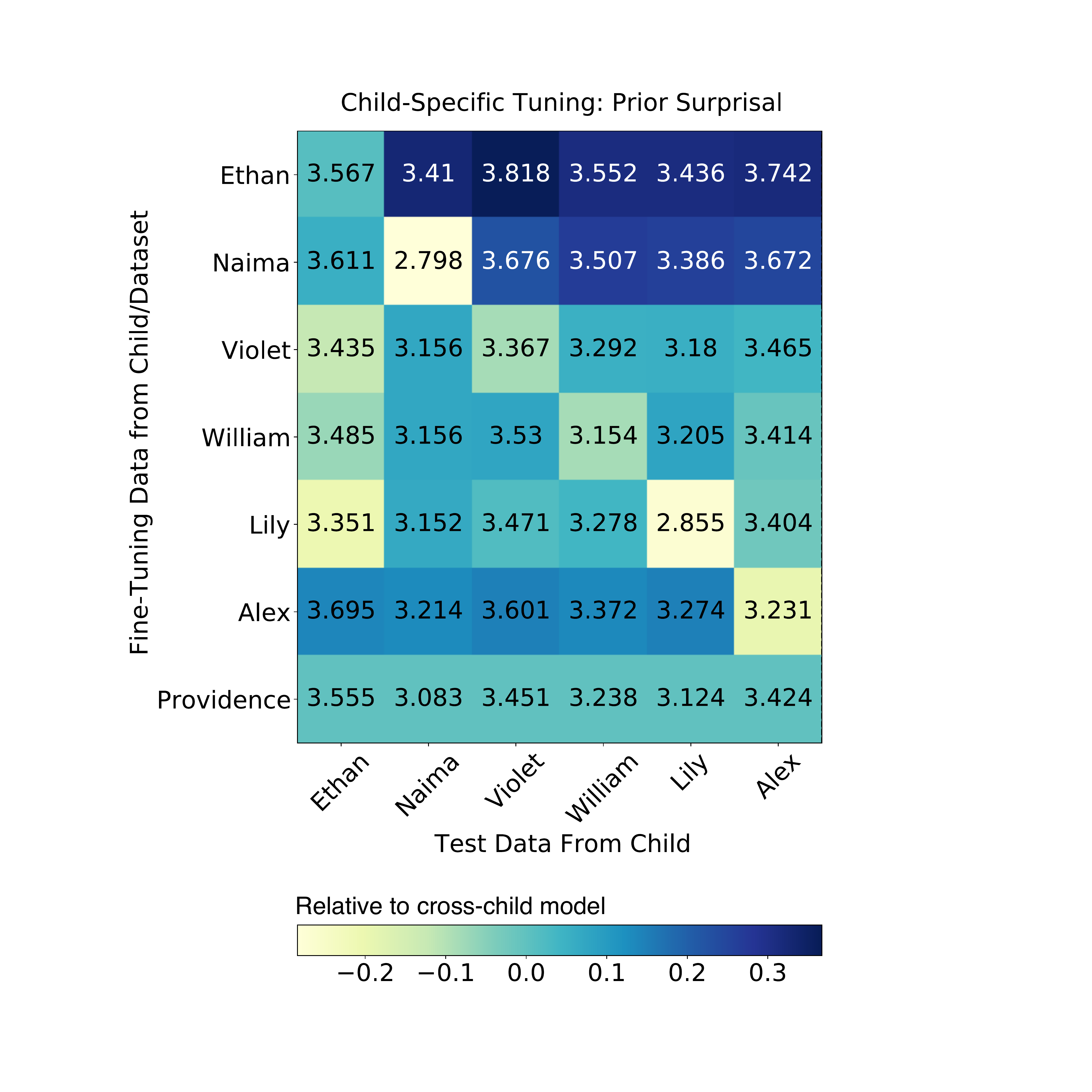}
 \put (-2,90) {\Large \textsf{A}}
    \end{overpic}
    \hspace{0em}
    \begin{overpic}[width=0.48\textwidth, trim=4cm 3cm 5cm 0cm,clip]{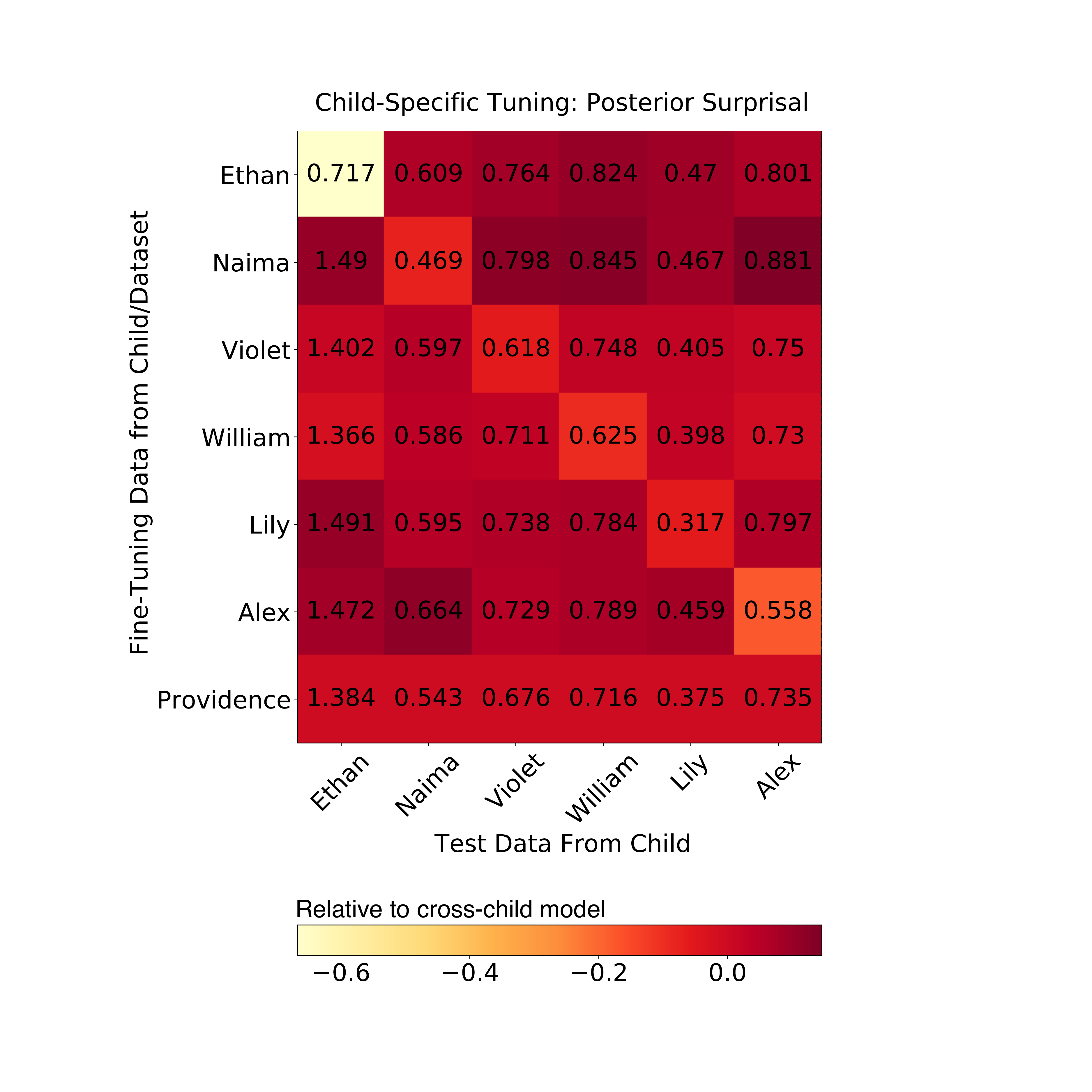}
 \put (-2,90) {\Large \textsf{B}}
    \end{overpic}
    \caption{\textbf{A.} Average prior surprisal and \textbf{B.} average posterior surprisal under the Phoneme-Specific likelihood for seven spoken word recognition models (lower is better).
    The first six models are fine-tuned on transcripts from specific children, while the remaining one is trained on a broader dataset: all children in the Providence corpus.
    Colors are normalized with respect to the  baseline performance of a model trained on data from all six children (final row); children are ordered by median age of their productions in the corpus.
    In A, the best-scoring prior is the one fine-tuned on other transcripts from that specific child for five of six children. 
    This suggests that adults---even transcribers---adapt to the constructions, topics and lexical choices of specific children.
    When combined with likelihoods tuned to each specific child (B), models fine-tuned to specific children always outperform those fine-tuned to other children on each child's test data.
    This provides evidence that adults' interpretations involve adapting to both the linguistic content and pronunciations of specific children.  
    }
    
    \label{fig:child_cross_wfst_posterior}
\end{figure*}

Besides tuning to the linguistic input of children of a specific age, an adult may also become attuned to the particular topics, constructions, and words that a specific child prefers.
Our modeling results shows strong evidence for this hypothesis:  the adult interpretation data for each child are better predicted by a model fine-tuned on the production data for that child than by models fine-tuned on the production data for any other child for five of six children (Fig. \ref{fig:child_cross_wfst_posterior}A, $p$=.002 by Monte Carlo simulation).\footnote{Scores for this analysis reflect a different test set than those in Table \ref{tab:model_surprisal_comparison}, and cannot be compared directly (see Methods).}
When combined with child-specific likelihood functions, posterior surprisal values are lowest for child-specific word recognition models in all six cases (Fig. \ref{fig:child_cross_wfst_posterior}; $p <$ .0001 by Monte Carlo simulation).
Taken together, these two analyses of age- and child- specific models demonstrate that child-directed listening models can be made yet more representative of adult inferential behaviors by further fine-tuning to specific children and developmental time periods. 
It also provides evidence that adults are themselves adapting to these same features when listening to children, as their interpretations are less likely under the models fit across all children.


\begin{table*}[h]
\centering
\tablefont
\caption{Example top guesses for the identity of the final word when a child says ``I want to [wid]'' (\textit{I want to weed}). Adult annotators labeled this word as ``read.'' Parenthetical indicates probability of top guess. Posterior probabilities reflect phoneme-specific likelihood.
\label{read_completions}}
\vspace{3mm}
\footnotesize
\begin{tabular}{l|cc|}
                                & \multicolumn{2}{l|}{Highest Probability Guess From...} \\ \cline{2-3} 
\textbf{Model}                           & \multicolumn{1}{c|}{\textbf{Prior}}              & \multicolumn{1}{c|}{\textbf{Posterior}}    \\ \hline
\textsc{BERT+CHILDES+Bidirectional}      & \multicolumn{1}{c|}{read (0.71)}        & read (1.0)   \\
\textsc{BERT+CHILDES+Preceding}          & \multicolumn{1}{c|}{read (0.74)}        & read (1.0)   \\
\textsc{BERT+Switchboard+Bidirectional} & \multicolumn{1}{c|}{read (0.57)}        & read (1.0)   \\
\textsc{GPT2+CHILDES+Preceding}          & \multicolumn{1}{c|}{read (0.61)}        & read (1.0)   \\
\textsc{Trigram+CHILDES}                 & \multicolumn{1}{c|}{go (0.1)}           & read (0.86)  \\
\textsc{Unigram+CHILDES}                 & \multicolumn{1}{c|}{i (0.04)}           & read (0.4)   \\
\textsc{UniformPrior}                    & \multicolumn{1}{c|}{All guesses equal}  & weed (0.53) 
\end{tabular}
\end{table*}

\section*{Discussion}

Here we find strong evidence of ``child-directed listening:'' sophisticated expectations on the part of adults regarding what messages children are most likely to communicate in early speech, as well as child-specific expectations for pronunciations.
Returning to the example of ``ah wan du weed'' we introduced at the outset of the paper,  Fig. \ref{read_completions} illustrates differences in the top guesses under the different models for the target vocalization ``weed.''
Bayesian cognitive models of spoken word recognition that use sophisticated neural-network--based priors can capture these implicit beliefs and produce interpretations of children's early language that are very similar to adults.
By contrast, simpler ablated models---such as models of word recognition that rely on knowledge of word sequence probability (\textsc{Trigram+CHILDES}), word probability (\textsc{Unigram+CHILDES}), or only the phonetic data produced by the child (\textsc{UniformPrior}) ---place lower probabilities on the adult interpretation.
This illustrates the richness of the inferences made by adults in the process of early communication, and points to ways in which adult cognition may facilitate communication between young children and adults.

While the current results do not speak directly to whether children learn more effectively as a consequence of child-directed listening, we outline below how this ability may have substantial implications for the nature of the learning challenge faced by children.
We then discuss the implications of the current work for methods of evaluating children's linguistic knowledge, and note several limitations.

First and foremost, these results invite a reconsideration of the nature of feedback in early language development \citep{chouinard2003adult, marcus1993negative, demetras1986feedback}.
For example, if we assume that successful communication is itself reinforcing, child-directed listening constitutes feedback to the child learner even in the absence of child-directed speech.
For example, a caregiver who interprets a child's production of ``uh'' to mean ``up'' may not say anything in response to the child's production, but may be providing a different form of feedback when they pick the child up.
This, in turn, leads to new puzzles: if adult caregivers can help many otherwise deficient attempts at communication succeed, what pressures children to get better?

Second, this deeper understanding of the adult helps to clarify how children's earliest language productions can serve as speech acts---linguistic productions that effect change in or via the listener---even as their knowledge of adult-like linguistic structure is still emerging (\eg \citep{dore1975holophrases}).
In this way, child-directed listening might provide scaffolding to support the gradual emergence of linguistic structure in support of communication.
The adult expectations described here are most useful for inferring word identities from noisy phonetic productions  when the range of expected messages from the child is most restricted, \ie in early development, when adults expect young children to use only a few words.
Adult expectations are less useful as children begin to use language to express a greater range of messages. 
Instead, communication will increasingly depend on the child's ability to distinctively articulate a growing inventory of words and multi-word constructions.

A serious consideration of child-directed listening has several implications for our most common tools for assessing child language: parental reports of vocabulary and transcribed speech. 
Children's articulatory maturity and vocabulary size are often evaluated through parent report (\eg Communicative Development Inventories, \citep{fenson2007macarthur}), but caregiver biases have long been recognized as a potential confound in the yielded data \citep{frank2021variability}.
The current work suggests that such biases may emerge as a natural consequence of adult inferential processes: for example, adults may posit the presence of nouns in ambiguous segments of children's speech by virtue of their prior expectations and the salience of contextual cues; thus, adults may be more likely to over-report nouns in their children's vocabularies. 
The current work may be extended into quantitative methods to gauge and correct for such biases.
For transcribed speech, the current work makes it clear that the interpretation of a child's word is highly dependent on cognitive processes in listeners.


We note a notable, yet arguably temporary, limitation with the current work: 
We make the simplifying assumption that inferences made by adult transcribers in the lab are representative of the inferences made by adult caregivers in the moment.
Transcribers have less shared history and only limited access to the broader non-linguistic context relative to adult caregivers who are interpreting their children's speech in realtime.
However, transcribers have substantial exposure to child language (often from the same child), along with opportunities to replay the child's speech, which caregivers do not have. 
We also note that transcribers may use speech and actions on the part of adult caregivers in the dataset to inform their interpretations of children's speech (\eg the parent responding to ``ah wan du weed'' with ``let's find a book!'').
In such cases, transcribers' interpretations may be especially representative of the inferences made by caregivers.
Nonetheless, potential differences in the inferential capacities of caregivers relative to other adult ``listeners'' should be tested experimentally.

\section*{Conclusion}
We present a suite of Bayesian models of spoken word recognition to test for and characterize \textit{child-directed listening}, or how adult caregivers find words in the noisy and often non-conventional speech productions of young children.
We find strong evidence that context-specific beliefs about what children are likely to say are critical for replicating adult interpretations of noisy child speech.
We further find evidence of child-specific adaptation: models that are fine-tuned to the pronunciations, topics, constructions, and word choices of specific children yield even better approximations of adult interpretations of those children's speech.
This research paves the way for new avenues of inquiry into how children become mature language users through the contributions of adult listeners.

\newpage
\section*{Methods}

\subsection*{Model Setup}

We treat the challenge of understanding what children mean---here, inferring the identity of a word used by a particular child in a particular linguistic context---as an instance of Bayesian inference (Fig. \ref{fig:overview_figure}). 
In our characterization, an adult listener seeks to recover a conventional word from the phonetic form produced by the child, \eg finding \textit{read} in the child's production ``weed'' (phoneme string [wid]). 
To combine the contributions of adult expectation with the phoneme string produced by the child, we adopt a Bayesian model of spoken word recognition (in the vein of \citep{norrisMcQueen2008}; see also \citep{norrisMcQueenCutler2016}). 
The model assigns a posterior probability to each candidate word interpretation $w$, taking into account the perceptual input data $d$ (\ie the child's phoneme string), and the linguistic context $c$:

\begin{align}
\label{eq:bayesian_swr} 
P(w|~d,c)  = \frac{P(d|w,c) P(w|c)} {\sum_{w'\in V}{P(d|w',c) P(w'|c)}}.
\end{align}


This formulation cashes out the intuition that the degree of belief that the child said word $w$  reflects the combination of 
\begin{inparaenum}[(a)]
    \item fit to perceptual data produced by the child and
    \item the listener's linguistic and non-linguistic expectations.
\end{inparaenum} 
``Fit to perceptual data'' is evaluated via the likelihood function, $P(d|w,c)$. 
For the current work, we assume that adults take a child's pronunciation of a word to be  independent of the broader linguistic context in which it appears, \ie we approximate $P(d|w,c) \approx P(d|w)$, though this assumption could be relaxed in future work.

The listener's expectations are captured by the prior probability $P(w|c)$ of the word $w$ in context $c$ (before perceptual input is taken into account).
Here we focus on the linguistic context, though we note that \textit{non}-linguistic information could be leveraged here as well (\eg people and salient objects visible in the scene; see also \citep{levy2008Expectation}).
The denominator in Equation~\ref{eq:bayesian_swr} reflects the summed strength of all competitor words $w'$ in the  vocabulary $V$.
Thus, the model guesses upon seeing \textit{both} the phonetic data and the surrounding context (technically a Bayesian \textit{posterior}) constitute a probability distribution over candidate words, with favored word interpretations receiving higher probabilities than disfavored ones.
In what follows, we discuss the likelihoods and then the priors for the set of models under consideration.

\subsection*{Likelihood} 
For the likelihood $P(d|w)$, or word pronunciation module, we tested three distinct likelihood functions.
We describe the simplest one first to develop intuitions regarding model behavior. 
In the simplest Edit-Distance likelihood, we use a transformation of string edit distance between the phoneme sequence for each candidate word interpretation and the phoneme string produced by the child.
Specifically, we use exponentiated negative edit distance \citep{levy2008noisy}:

\begin{align}
\label{eq:bayesian_swr_likelihood2} 
P(d|w) \propto e ^{-\beta~\times~ \text{dist}(d^{*}:w,~d)}
\end{align}

\noindent where $dist$ is the edit distance (or Levenshtein distance, or the minimal number of deletions, insertions and substitutions) between citation form $d^*$ for candidate word $w$, designated here $(d^{*}:w)$, and the observed child phonetic form $(d)$.
If a word $w$ has more than one citation form, \eg \textit{read} can be pronounced as [\textipa{\*r}id] or as [\textipa{\*rE}d], we iterate over all pronunciations and take the smallest distance.
The free parameter $\beta$ scales these likelihoods: as $\beta$ approaches 0, all likelihoods approach 1.

This treatment of edit distance does not take into account phoneme similarity, \ie that certain phonemes are more perceptually similar, that children are more likely to make certain substitutions or deletions compared to others  (\eg, that [wid] is a relatively likely child pronunciation for \textit{read}, [\textipa{\*r}id]), or that certain edits are more likely in combination. 
This means that the Edit-Distance likelihood may not fully capture adults' abilities to use the phonetic form to inform their interpretation.
For this reason, we include a more sophisticated Phoneme-Specific model of child pronunciation, in the form of a \textit{weighted finite state string transducer} \citep{mohri2002weighted, gorman2020sigmorphon}.
A transducer is a probabilistic or deterministic mapping between strings in an input language and strings in an output language. 

Here, we train the mapping on pairs of citation forms for words coupled with children's pronunciations from the Providence corpus.
Following \citep{novak2016phonetisaurus}, we learn an alignment between adult and child pronunciation through expectation maximization \citep{dempster1977maximum} using the \texttt{BaumWelch} library \citep{gormanBaumWelch2021}.
We use a learning rate of 1.0, representing weights in the tropical semiring, and normalize the weights of all arcs leaving a state with the same input label sum to one. 
We then learn a \textit{joint} unigram model (\citep{galescu2001bi}; see also pair language models, \citep{novak2012wfst}) over the aligned pairs (\ie each event in the $n$-gram model is a combination of an input symbol and an output symbol).
With this phoneme-specific model, we can then estimate the probability that each possible candidate word $w$ would generate the child's observed pronunciation $d$.

The transducer yielded by the training procedure in the main text encodes joint probabilities of edits (input-output pairs) on strings.  
Our goal then is to estimate the probability of a given output string (the phonetic form produced by the child) given an input string (the conventional adult form), similar to the Edit-Distance likelihood above.
Thus we adopt a similar likelihood function to the Levenshtein distance one above,

\vspace{-3mm}
\begin{align}
\label{eq:bayesian_swr_wfst_likelihood} 
P(d|w) \propto e ^{-\lambda~\times~ \theta_{\text{WFST}}(d^{*}:w,~d)}.
\end{align}
\vspace{5mm}

Whereas the Edit-Distance likelihood relies on a single minimum edit distance between the input and output string, the Phoneme-Specific likelihood relies instead on the aggregate weight of paths through a finite state machine, $\theta_{\text{WFST}}(d^*:w,~d)$, which captures the probabilities of the different edit sequences that would transform the adult conventional form into the child's produced form. 
We create a finite state machine, $\text{FSM}(d^*,d)$, for each combination of 7,997 word identities in the candidate vocabulary and phonetic form produced by the child.
Computationally, this involves composing a finite state acceptor for the conventional adult form $d^*$ for each possible word identity with the transducer learned above, and then composing that machine with the finite state acceptor for the child's produced phonetic form, $d$. 
We then enumerate the weights along all paths through the finite state machine,

\begin{align}
\label{eq:sum_path_probs} 
\theta_{\text{WFST}}(d^*:w,~d) = -1 \times \log \left( \sum_{\text{path} \in \text{FSM}(d^*,d)}{P(\text{path})}\right).
\end{align}

\noindent Following the chain rule, each path reflects the product of the conditional probabilities of its component arcs:

\begin{align}
\label{eq:product_of_arcs} 
P(\text{path}) = \prod_{\text{arc} \in \text{path}}{P_{\text{FSM}(d^*,d)}(\text{arc})}.
\end{align}

\noindent To obtain the \textit{conditional} probabilities appropriate for this computation (rather than a joint probability, as yielded by expectation maximization) we normalize the arc probabilities the probability of all \textit{possible} outputs given the input (all arcs from a given state),

\begin{align}
\label{eq:joint_ngram_no_history} 
P_{\text{FSM}(d^*,d)}(\text{arc}) =  \frac{P(d^*_i \rightarrow d_i)}{\sum_{s \in \mathbb{S}}{P(d^*_i \rightarrow s)}},
\end{align}

\noindent where $s$ indicates each possible output symbol in the symbol set $\mathbb{S}$. 
In the unigram case, the probability of edit $d^*_i \rightarrow d_i$ is not conditioned on any history, thus we simply divide by the summed probability of all possible edits where $d_i$ is the input symbol. 
Note that the Phoneme-Specific likelihood would yield probabilities proportional to the simpler Edit-Distance likelihood if the probability (\ie cost) of all edits is equal to 1 in Eq. \ref{eq:joint_ngram_no_history} and only the lowest cost path were considered.
As with the simpler Edit-Distance likelihood, we iterate over all pronunciations for a given orthographic word and take the path weights associated with the citation form that assigns the highest probability to the observed data.

For the Edit-Distance likelihood, we evaluated posterior surprisal estimates for values of $\beta$ between 1.5 and 4.5 by $0.1$ increments, and take the value that assigns the highest posterior probability to a sample of 5,000 transcriber word interpretations. 
We confirmed that the resulting scores were convex across the parameter range and that the highest scoring parameter value was not on the edge of the range of possible values.
In the case of the Phoneme-Specific likelihood, we grid sample the scaling parameter $\lambda$, analogous to $\beta$ for the Edit-Distance likelihood, although we focus on the range 0-2 by $0.1$ increments.

The third likelihood function we test (Expt. 3) reflects the same approach as the Phoneme-Specific likelihood, however we fit a separate WFST for each child, and follow the an analogous fitting procedure.
This captures the intuition that listeners may learn how individual children are likely to deviate from pronunciations in adult speech.
Given the primary goal of evaluating adults' prior expectations when listening to children, we test child-specific likelihoods only in combination with child-specific priors.

\subsection*{Priors: Language Models} 
For each child vocalization, we retrieve prior probabilities over candidate word interpretations using a suite of probabilistic language models. 
As a ``best'' prior architecture capable of capturing long-distance syntactic dependencies, discourse patterns, and child-specific constructions we use \textsc{BERT} \citep{devlin2018}, which has demonstrated extremely competitive performance for single-word completion tasks, including automatic speech recognition \citep{salazarEtAl2020}.
By virtue of its attentional mechanisms, BERT is able to effectively model long distance dependencies \citep{jawahar2019}, and capture speech register and discourse-level information.
We compute the probabilities for the masked word $P(w|c)$ from BERT, using a language modeling head with the \texttt{transformers} library \citep{wolfEtAl2020}.
We treat the identities of words \textit{other than the masked word} as sufficiently reliable to predict the identity of the masked position in the utterance (cf. \citep{salazarEtAl2020}). 
For each masked phoneme sequence, we take the real-valued vector of predictions corresponding to the model's full vocabulary, extract the activations corresponding to the candidate vocabulary above, and compute the softmax to yield a vector of probabilities over each of the candidate words in the vocabulary.

We also test another transformer-based model, \textsc{GPT-2}, which uses a similar attentional mechanism to BERT but predicts word identities sequentially (\ie left-to-right in an utterance). 
Like BERT, GPT-2 can effectively model long-distance dependencies and discourse topics.
In the same fashion as BERT, we take a real-valued vector of activations corresponding to the candidate vocabulary and compute the softmax (see Tokenization below regarding how we achieved parity in vocabularies between GPT-2 and BERT).

\subsubsection*{Fine Tuning}
For fine tuning BERT, we initialized a standard \texttt{BERT-uncased} model using the \texttt{transfomers} library, then updated the weights to best predict the identities of words in a new training set --- in this case, utterances from 80\% of North American English and UK English CHILDES transcripts.
This dataset consists of 4,119,182 utterances and 15,885,051 words (an additional 20\% of transcripts were held out for validation).
We followed a similar procedure to finetune GPT-2 models.
These fine-tuned models (\textsc{BERT+CHILDES+Bidirectional}, \textsc{BERT+CHILDES+Preceding}, \textsc{BERT+CHILDES+OneUtt}, \textsc{GPT2+CHILDES+Preceding}, \textsc{GPT2+CHILDES+OneUtt}) should be expected to be more representative of adult linguistic expectations in understanding child speech than the standard BERT and GPT-2 models respectively for three reasons.
First, they should assign higher probability to words that are common in speech to and from children.
Second, they should assign higher probability to non-sentence fragments, which are ubiquitous in conversational speech but somewhat less prevalent in adult-directed written language. 
Third, they should encode an expectation for the dyadic, back-and-forth structure of scenes typically captured in transcripts (see Speaker Identity, below).

To test for the advantages of fine-tuning on utterances from the child language environment, we fine-tune two more BERT models (\textsc{BERT+Switchboard+Bidirectional}, \textsc{BERT+Switchboard+OneUtt}) using a large adult-to-adult conversational speech corpus, the Switchboard corpus \citep{godfrey1992switchboard}.
This training sample includes 95,786 utterances and 1,175,384 words in the training sample. 
While we expect that fine-tuning on a conversational model will produce a model that expects shorter utterances, a higher proportion of sentence fragments, and more filled pauses, it does not contain the conversational topics and constructions preferred by children.

\subsubsection*{Scope of Linguistic Context} In addition to fine-tuning the model, we manipulate whether prior estimates reflect access to the larger discourse context as captured by the transcript before and after a particular vocalization. 
Because these models are meant to be representative of \textit{caregiver} expectations, these models condition their predictions regarding word identity on what the caregiver and child \textit{both} say, both before and after the masked token.
In \textsc{+Bidirectional} models, we allow the model to see 20 utterances preceding and following each mask during inference; while in \textsc{+Preceding} the model sees only the 20 preceding utterances plus the utterance containing the target vocalization.
For GPT-2 models, \textsc{+Preceding} refers to strictly words preceding the target vocalization.
For BERT models, \textsc{+Preceding} refers to those preceding the target vocalization as well as those in the same utterance because BERT is not suited to computing probabilities of continuations. 

\subsubsection*{Speaker Identity} One option for fine-tuning models is to include speaker identities as tokens (\eg \texttt{[CGV]} for caregiver and \texttt{[CHI]} for child) and include them in the fine-tuning process.
This allows the model to condition its predictions on the current speaker, for example providing different completions for ``[CHI] I want to [MASK]'' vs. ``[CGV] I want to [MASK].''
Both fine-tuned BERT and GPT-2 models with speaker information uniformly outperformed ones without, so for brevity we report on only these in the current paper.
Both Switchboard and standard BERT models are trained on written data that lack speaker identifiers, so this information cannot be used in inference for these models. 

\subsubsection*{Tokenization}

The standard BERT model has its own vocabulary, which imposes limitations on the vocabulary in the analysis.
Standard implementations of BERT split longer words into ``word pieces,'' or most common repeated sub-sequences.
In English, this often yields an approximation of morphological segmentation (\eg \textit{fishing} $\rightarrow$ \texttt{fish}~~\texttt{\#\#ing}), but many word pieces are not morphemes (unless the model is specifically trained to yield such, \eg \citep{hofmann-etal-2021-superbizarre}).
For the purposes of predicting a single masked word, BERT predicts only one word piece. 
We thus limit the vocabulary to word-initial word pieces like \textit{fish}, and exclude continuations like \texttt{\#\#ing} from consideration.\footnote{Hence we do \textit{not} consider each phonetic sequence as a possible continuation of the preceding word, though this is logically possible.}
This means that the model largely excludes morphologically-rich words, though children relatively rarely produce such words during this age range. Further, we expect that the general approach of child-directed listening holds for these words as well.

GPT-2 uses a different tokenization scheme, called byte pair encoding \citep{shibata1999byte}, such that the set of tokens predicted by the model differs from the BERT models above.
To achieve parity in the vocabularies across these models, we added any words in the vocabulary of the BERT models to the tokenizer and learned representations for these words during the fine-tuning process.

\subsubsection*{Age and Child-Specific Fine Tuning}

For age-specific fine-tuning, we fine-tuned separate BERT models on language samples from the CHILDES dataset from children at or below 30 months, vs. above.
20\% of transcripts were held out for model validation.
We then evaluated these models on the same test set as all other models, such that the results in Fig. \ref{fig:posterior_surprisal_by_age} are comparable to the scores in Tab. \ref{tab:model_surprisal_comparison}.

The child-specific fine-tuning analysis required a careful approach to clearly separate training, validation, and test data from the Providence corpus in order to avoid overfitting (\ie testing a model on words it saw during training).
We thus created for each child test and validation sets consisting of an age-balanced selection of randomly selected transcripts and apportioned the remainder to the training set for that child.
This age stratification strategy helps to avoid artifacts in performance that might emerge from differences in ages in the composition of test data (\ie the model will appear to perform better when more test items come from when children are older).
We did not use an age-stratified training set out of a concern for data sparsity.
We used each training set to fine-tune a variant of the \textsc{BERT+CHILDES+Bidirectional} model for each child; we then also used all words with phonetic transcripts from that training set to fit a child-specific WFST to use in the model likelihood.  
We then fit a free parameter in each likelihood ($\beta$ for the Edit-Distancelikelihood and $\lambda$ for the Phoneme-Specific one) separately for every combination of train and test dataset (\ie each cell in Fig. \ref{fig:child_cross_wfst_posterior}B).

\subsubsection*{Ablated Models}
To evaluate a baseline where adults solely use the phonetic input supplied by the child, we include a \textsc{UniformPrior} model where all English mono- and bi-syllabic words in the child language environment are equally likely \textit{a priori} and the model must rely on the perceptual input to identify the child's intended word.
This model assigns equal probability to all words ($1/|V|$, where $|V|$is the number of candidates). 
This provides the comparison case of a maximally uninformative prior.

We include a unigram, or frequency-based model (\textsc{Unigram+CHILDES}) which uses the normalized frequencies of words in the same dataset used to fine-tune BERT and GPT models (see Fine Tuning, above). 
To avoid assigning zero probability to any unseen words (\ie words in the Providence corpus but not in CHILDES), we add a small pseudocount (.001) smoothing to all counts before computing probabilities.
This model reflects the hypothesis that adults expect young children to use specific words, but do not expect them to use adult-like syntactic structures or adhere to common patterns of discourse, such as alternating turn-based contributions.

We also test an $n$-gram model trained on the same large sample of child language as the fine-tuned neural network models.
This trigram model is more sophisticated than the unigram model in that it takes into account sequential dependencies among words, but is less sophisticated than the neural network models because it cannot track long distance dependencies.
We employ Kneser-Ney smoothing \citep{chen1999empirical} to re-allocate probability mass to unseen bigrams and trigrams on the basis of commonly observed unigrams and bigrams, respectively.
We query this trigram model in two different ways in the word prediction module. 
In \textsc{Trigram+CHILDES}, we extract the probability distribution over continuations in the vocabulary using up to the preceding two words in the utterance, $P(w_{i}|w_{i-2}, w_{i-1})$.
In the second method, \textsc{Trigram+CHILDES+OneUtt}, we compute the probability of the resulting utterance for each item in the vocabulary under the $n$-gram model, testing probability in context:


\begin{align}
\label{eq:ngram_conditional} 
P(w_i) = \frac{P(w_{<i}) P(w_i|w_{<i}) P(w_{>i} | w_{<i},w_i)}
{\sum_{w'_i} P(w_{<i}) P(w'_i|w_{<i}) P(w_{>i}|w_{<i},w'_i)} 
\end{align}.

\noindent This evaluates the probability of $w_i$ as a continuation of preceding words through the second term, $P(w_i|w_{<i})$, as well as the probability of the resulting continuations through the third term,  $P(w_{>i} | w_{<i},w_i)$. This is normalized by the probability of all possible words in the vocabulary (iterating through $w'$ in the denominator).

\subsection*{Data Retrieval and Processing}
Utterances from both parents and children and phonological transcripts from children from the Providence corpus \citep{demuthEtAl2006, demuth2009prosodic} were retrieved through childes-db 2020.1 \citep{sanchez2019}. 
The dataset contains broad phonetic transcriptions by trained transcribers for a subset of children's speech in the Phonbank format \citep{rose2014phonBank}. 
Wherever transcribers identified English words in children's speech, transcripts were seeded with phonetic transcriptions with adult citation forms from the CMU pronunciation dictionary (henceforth CMU dictionary); thereafter a separate group of phonology-focused transcribers edited the transcriptions word-by-word to reflect children's actual productions.
Other phonetic material was transcribed directly.
Further details regarding participants, coding procedure, and evaluations of inter-annotator agreement can be found in \citep{demuth2009prosodic}.

For the current work, we identified instances where adult transcribers assigned a conventional American English interpretation to a child's vocalizations. 
More specifically, we identified tokens produced by children in the intersection of four criteria:  
\begin{inparaenum}[(1)]
    \item possessing mono- or bi-syllabic phonetic forms (motivated below in \textit{Limiting to One and Two-Syllable Vocalizations})
    \item possessing \textit{no} unintelligible (CHILDES code \texttt{xxx}) or phonology-only  (\texttt{yyy}) tokens in the same utterance \footnote{Handling multiple unknown tokens per utterance is entirely possible, but would require a more sophisticated modeling approach and significantly more computational resources than the one presented here, as the model would need to consider all possible interpretations of utterances. Here, we limit ourselves to the simpler model of individual word recognition for accessibility to a broader audience.}
    \item whose gloss is present as a token in BERT (motivated below in \textit{Tokenization})  %
    \item whose gloss is included in the CMU pronouncing Dictionary.
\end{inparaenum}

We also identified instances where the transcribers did not assign a conventional English interpretation.
These instances had to meet the first criterion above, but received the special gloss code of \texttt{yyy}.
This code indicates that the vocalization could receive a phonetic transcript, but not an adult English interpretation (compare with Phonbank's \texttt{xxx} code, indicating that a vocalization was unintelligible in both respects, often as a consequence of environmental noise).
Following best practices from machine learning for ensuring generalizability to new data, we split the dataset into  \texttt{validation} and \texttt{evaluation} partitions and developed model architectures with respect to validation, and ran our models on samples in the test partition \textit{after} committing to the model architectures and fitting procedures. 

\subsection*{Limiting to One and Two-Syllable Vocalizations}

We restricted our analysis to one- and two- syllable vocalizations produced by the child. 
Vocalizations with three or more syllables are problematic for two reasons.
First, they are much more likely to contain multiple distinct words, which would require considering multi-word sequences as possible interpretations of the phonetic signal produced by the child, which would in turn require a more complex modeling approach.
Second, even if a longer vocalization corresponds to a single word, it is much less likely to be present as a separate token in the model vocabularies, meaning that the model would still need to consider multi-token sequences (\eg sequences of ``word piece'' tokens in BERT) as interpretations. 
While we could easily articulate such a model, it would add considerably to the complexity of the proposed model and be much more computationally intensive.
For this reason, we restrict our vocabulary as well as test items to short vocalizations.
Syllables were counted by enumerating the number of vowel nuclei separated by consonants.

\subsection*{Candidate Vocabulary} 
The vocabulary considered by each model was the intersection of 
\begin{inparaenum}[(1)]
    \item{words in the CMU dictionary with one or two syllables and}
    \item{tokens present in BERT  (motivated below)}
    \item{tokens that appeared four or more times in CHILDES (to limit to words that might reasonably be said by English-learning children).}
\end{inparaenum}
The final vocabulary included 7,997 orthographic words.
Each orthographic word could map to one or more pronunciations, for example that \textit{read} could be pronounced as [\textipa{\*r}id] or [\textipa{\*r}\textipa{E}d], and that \textit{red} could also be pronounced as [\textipa{\*r}\textipa{E}d].
In total we considered 8,943 possible pronunciations.
We reconciled differences in IPA conventions between the Providence Corpus and the CMU dictionary following a procedure detailed in our codebase (\url{osf.io/v7c3e/}).

\subsection*{Evaluation}
Our evaluation tests whether our models can predict what adults interpreted children as saying in this ``guess the word'' setup. 
More specifically, we take the posterior probabilities of word identities yielded by each of these 26 models (thirteen priors $\times$ two likelihoods) and test how well they predict adult interpretations of child phoneme strings.
Ideally, we would use a large dataset of caregivers' realtime inferences of what their children said.
However, such a large datasets would be prohibitively difficult to collect. 
Instead, we use the fact that annotators who create transcripts of child language datasets are faced with a similar inferential problem: assigning interpretations to children's speech. 
Our test corpus (the Providence corpus, covering $n$ = 6 children, 11 -- 48 months old, and containing 460,000 utterances; \citep{demuthEtAl2006, demuth2009prosodic}) contains matched pairs of phonetic transcriptions and English adult word interpretations provided by trained annotators.

\begin{table}
\centering
\tablefont
\begin{threeparttable}
\caption{Example model inputs and guesses for four of the sixteen Bayesian spoken word recognition models we test. 
Each model receives a phoneme sequence and the surrounding utterance or utterances  (A, B) and produces a posterior probability distribution over possible word interpretations for what the word indicated by the empty square may be.
C. shows the highest prior probability guesses (before observing the data) corresponding to the different probabilistic language models used as priors.
Each models' top 4 guesses are shown ordered left to right by their model-estimated probability; the word recovered by the transcriber is indicated with italics (only on left side, A, since no word was recovered by transcribers on the right, B).
D shows model guesses from each model's posterior probability distribution, which reflect both the prior expectations from C as well as the fit to the child-produced data (i.e. based on the phonetic input from the child). 
This basic setup can be used both in cases where transcribers interpret children's speech as a word (A) or treat it as unintelligible (B). \label{example_table}}
\setlength{\tabcolsep}{0.3em} 
{\renewcommand{\arraystretch}{1}
\begin{tabular}{lllcll}
\toprule
\multirow{7}{*}{\textbf{Inputs to Model}} & \multicolumn{2}{c}{\textbf{A. Transcriber Finds a Word (\textit{read})}} 
    &~& \multicolumn{2}{c}{\textbf{B. Transcriber Labels Form as Unintelligible}}\\[.25ex]            
\cline{2-3} \cline{5-6}\\[-1.5ex] 

&\textcolor{gray}{Mother} & \textcolor{gray}{this is}                                             
    &~& 
        \textcolor{gray}{Mother} & \textcolor{gray}{do you want ta put some beans in your eggs?}             \\
& \textcolor{gray}{Mother} & \textcolor{gray}{you want mamma let's see}                                         
    &~& 
        \textcolor{gray}{Child} & \textcolor{gray}{no} \\[.5ex]
\multicolumn{1}{r}{} & Child & / {\textipa{A}\textipa{@}~~w\textipa{A}n~~d\textipa{@}~~wid} /  &~& 
    Child & / ju~~~~m\textipa{E}\textipa{I}k~~yo\textipa{@}\textipa{\*r}~~~f\textipa{E}t /\\
\multicolumn{1}{r}{}  && \textit{~~I~~~~want~~to~~\fbox{$~~~~~~^{~}$}}
    &~& 
         & \textit{~you~~make~~your}~\fbox{$~~~~~~^{~}$}       \\[.5ex]
& \textcolor{gray}{Mother} & \textcolor{gray}{okay that's fine}                                                   
    &~& 
        \textcolor{gray}{Mother} & \textcolor{gray}{can I make one?}                                      \\
& \textcolor{gray}{Mother} & \textcolor{gray}{okay mommy's gonna pick out a book}                     
    &~& 
        \textcolor{gray}{Mother} & \textcolor{gray}{no} \\[.5ex]
\midrule\\[-2ex]
\textbf{Language Model} & \multicolumn{5}{c}{\textbf{C. Highest \colorbox{black}{\textcolor{white}{\textsc{Prior}}} Probability Model Guesses for} \fbox{$~~~~~~^{~}$}} \\[.5ex] 
\cline{2-6}\\[-2ex]
\textsc{cdl+context}    & 
    \multicolumn{2}{l}{see (0.55)~~~~go (0.05)~~~~play (0.04)~~~\textit{read} (0.03)~~[...]}                                    
    &~& 
        \multicolumn{2}{l}{{own (0.16) house (0.05) bed (0.05) dinner (0.03)~~[...]}}            \\
\textsc{bert+context}      & 
    \multicolumn{2}{l}{\textit{read} (.49)~~~~see (.28)~~~~~play (.04)~~~~know (.04)~~[...]}                       
    &~& 
        \multicolumn{2}{l}{{own (.25)~~~~choice~~~~(.24)~~~~point (.04)~~~~bed (.03) call (.03)~~[...]}}                \\
\textsc{childes-1gram}            & 
    \multicolumn{2}{l}{I~~~(.04)~~~a (.03)~~~~the (.03)~~~yeah (.03)~~~no (0.03)~~[...]} 
    &~& 
        \multicolumn{2}{l}{I (.04)~~~~~~a (.03)~~~~~the (.03)~~~~~yeah (.03)~~~~~~no (0.03)~~[...]}
        \\
        \textsc{UniformPrior}            & 
    \multicolumn{2}{l}{\textit{~~~~~All word identities equiprobable \emph{a priori}}} 
        &~& 
            \multicolumn{2}{c}{\textit{All word identities equiprobable \emph{a priori}}}       \\[.5ex] 
\bottomrule\\[-2ex]
\textbf{Language Model} & \multicolumn{5}{c}{\textbf{D. Highest \colorbox{black}{\textcolor{white}{\textsc{Posterior}}} Probability Model Guesses for} \fbox{$~~~~~~^{~}$} \textbf{(using WFST Likelihood)}}  \\[.5ex]
\cline{2-6}\\[-2ex]
\textsc{cdl+context}    & 
    \multicolumn{2}{l}{\textit{read} (0.994)~~wait (0.001)~~weed (0.001)~~~~what (0.001)~~[...]}                                
    &~& 
        \multicolumn{2}{l}{favorite (0.35)~~~~first (0.21)~~~~fort (0.12)~~~~face (0.07)~~[...]}            \\
\textsc{bert+context}      & 
    \multicolumn{2}{l}{\textit{read} (0.49)~~~~see (0.28)~~~~play (0.04)~~~~know (0.04)~~[...]}                                     
    &~& 
        \multicolumn{2}{l}{favorite (0.34)~~~~first (0.26)~~~~effort (0.14)~~~~fort (0.06)~~[...]}         \\
\textsc{childes-1gram}            & 
    \multicolumn{2}{l}{
    \textit{read} (0.39)~~~~with (0.25)~~~~what (0.12)~~~~would (0.04)~~[...]} 
        &~& 
            \multicolumn{2}{l}{first (0.38)~~~~for (0.32)~~~~front (0.08)~~~~different (0.03)~~[...]}       \\
            \textsc{UniformPrior}           & 
    \multicolumn{2}{l}{weed (0.37)~~~~worried (0.17)~~~~reed (0.06)~~~~\textit{read} (0.06)~~[...]} 
        &~& 
            \multicolumn{2}{l}{freight (0.2)~~~freighter (0.09)~~~fort (0.09)~~~flirt (0.08)~~[...]}       \\
            [.5ex] 
\bottomrule\\[-2ex]

\end{tabular}
}
    
\end{threeparttable}
\end{table}



In Experiment 1, we test our models' ability to predict whether transcribers deemed a particular vocalization to be a word versus flagged it as unintelligible.
We divide our set of child phoneme strings into instances where transcribers posited a word interpretation  (Table \ref{example_table}A)  and instances where they did not (Table \ref{example_table}B).

We take the entropy of the model's posterior distribution over candidate words $w$, $P(w|d,c)$, as the key quantity predicting whether a given perceptual input token $d$ in context $c$ will be interpreted as a word or as unintelligible:

\vspace{-5mm}
\begin{align}
\label{eq:entropy_def} 
H(w|d,c) = -\sum_{i=1}^n{P(w|d,c)\log P(w|d,c)},
\end{align}
\vspace{-3mm}

\noindent 
We assume that the lower the entropy, the more likely $d$ will be deemed to be intelligible and assigned a word interpretation. 
We assess each model's performance by the area under its receiver operating characteristic curve (AUC; \citep{fawcett2006roc}).
We evaluated 57,812 phonetic sequences that transcribers assigned a word interpretation, and 12,786 instances where they did not.

In Experiment 2, we test these Bayesian spoken word recognition models in their ability to reproduce transcribers' word interpretations.
Focusing on only these cases where transcribers interpreted children's productions as transcribable words, we compute the surprisal \citep{levy2008noisy, hale2001probabilistic} or self-information that each model assigns to the transcriber's interpretation $w$,

\vspace{-5mm}
\begin{align}
\label{eq:surprisal_def} 
I(w|d,c) = -\log_{2} P_{model}(w | d,c),
\end{align}
\vspace{-3mm}

\noindent and then take the average across the 57,812 instances where transcribers found words in our dataset for each model.\footnote{\ie the per-instance log-likelihood of the data under each model. 
We adopt this transformation for ease of presentation, but caution against interpreting these numbers as representative of real-time difficulty of language processing \citep{levy2008noisy, hale2001probabilistic} given that many of our models can use both preceding \textit{and following} linguistic context to predict a word's identity.}
The lower the average surprisal of the transcriber's word interpretation, the better the model reflects adult inferences about child speech.
In addition, we report the average rank of the transcriber's word interpretations for each model (ideal model performance would result in a mean rank of 1, where the transcriber's word is always the highest ranked word interpretation according to the model).
We also report the percentage of correct top guesses, which captures how often the word found by the annotator is the same as the highest ranked interpretation under each model.

In Experiment 3, we investigate additional age-based and child-specific fine-tuning of the prior, as well as child-specific fine-tuning of the likelihood.
For the age-based priors, we compute the posterior surprisal averaged across a sample of child vocalizations per each 6-month age bin.  
To evaluate the child-specific models, we run each model on the six child-specific test sets.

\subsection*{Model and Data Availability}
All model training and analysis code, as well as the fine-tuned model pre-processed child transcripts can be accessed through our Open Science Foundation repository at \url{osf.io/v7c3e/}.


\section*{Acknowledgements}
We thank Jessica Mankewitz, Sathvik Nair, and Rachel Jansen for providing feedback on early drafts as well as members of the Computational Psycholinguistics Lab at MIT and the Bergelson Lab at Duke for valuable discussion. We thank Kyle Gorman, Tiwa Eisape, and Peng Qian for several helpful technical consultations. Sophia Zhi contributed to the implementation of the pronunciation module. This work was supported by NSF grants BCS-1551866, BCS-1844710, and BCS-2121074; NIH grant 1F32HD097982; and the Simons Center for the Social Brain.


\newpage
\bibliography{cdl-arxiv-main}

\clearpage

\section*{Supplementary Information}

\subsection*{Supplementary Note: Spoken Word Recognition Model Performance By Child}

\begin{figure*}[h]
\centering
\includegraphics[width=.6\linewidth]{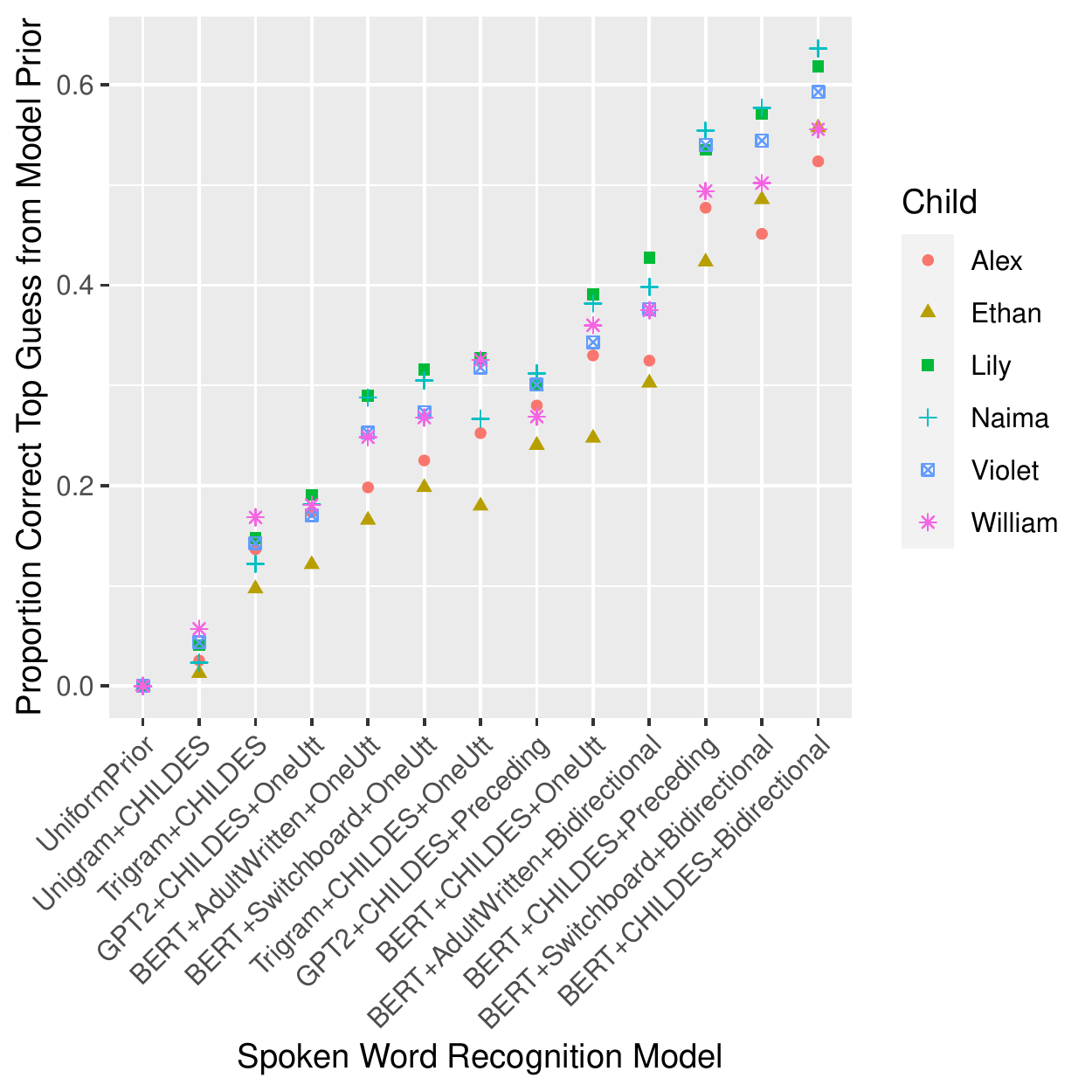}
\caption{
Proportion of instances where each model agrees with the annotator's word identity by child across the thirteen models used in Expts. 1 and 2
(analogous cross-child model results are presented in ``Prior Correct Top Guess'' in Table 2). 
Models are plotted left to right based on the model's average performance across children.
}
\label{fig:prior_by_child}
\end{figure*}

To what degree does performance of the spoken word recognition models used in Experiments 1 and 2 vary by child? 
Given that the word prediction module for each model is trained on language samples from children outside of the test set, does this general, cross-child model reproduce adult interpretations better for some children in the test set compared to others?
Here we re-analyze the results of Expt. 2 by calculating the proportion of top-1 guesses per each model (\ie how often the model's top guess was the word identity ascribed by an annotator) and splitting performance by child.
Fig. S\ref{fig:prior_by_child} shows model performance when predicting word identities based on prior expectations---\textit{before seeing data}---is relatively stable across children, with the model architecture / training dataset accounting for substantially more variance than the identity of the child.

Once the models observe the word form produced by the child and use the pronunciation module to produce a posterior update, the performance by child is more clearly distinguished (Lily $>$ Violet $>$ Naima $>$ Alex $>$ William $>$ Ethan). 
This suggests that both likelihoods tested for the word pronunciation module are more sensitive to the identity of the child: some children's articulations that are less noisy than others with respect to a cross-child model of pronunciation. 
We note that the Phoneme-Specific pronunciation module could, in principle, perform better on some children rather than others because more annotator labels + pronunciations came from some children rather than others.
However, we note that the same ordering of children exists for the models using the Levenshtein Distance pronunciation module, which reflects only how far the child pronunciations are from the standard adult pronunciation (\ie, not reflecting the amount of data from each child used to train the model).
This suggests that performance is largely dependent on how similar children's productions are to word pronunciations in the adult language.

We note that differences in performance across children in both cases (model priors and model posteriors) could arise from differences in the age coverage of the test set: the models perform better on speech from older children (Figure 5), and some children may have more test items from older ages.
We more rigorously approach this question of cross-child variability --- how to equitably compare it and how adults might adapt to it --- in Expt. 3.

\begin{figure*}[h]
\centering
\includegraphics[width=.99\linewidth]{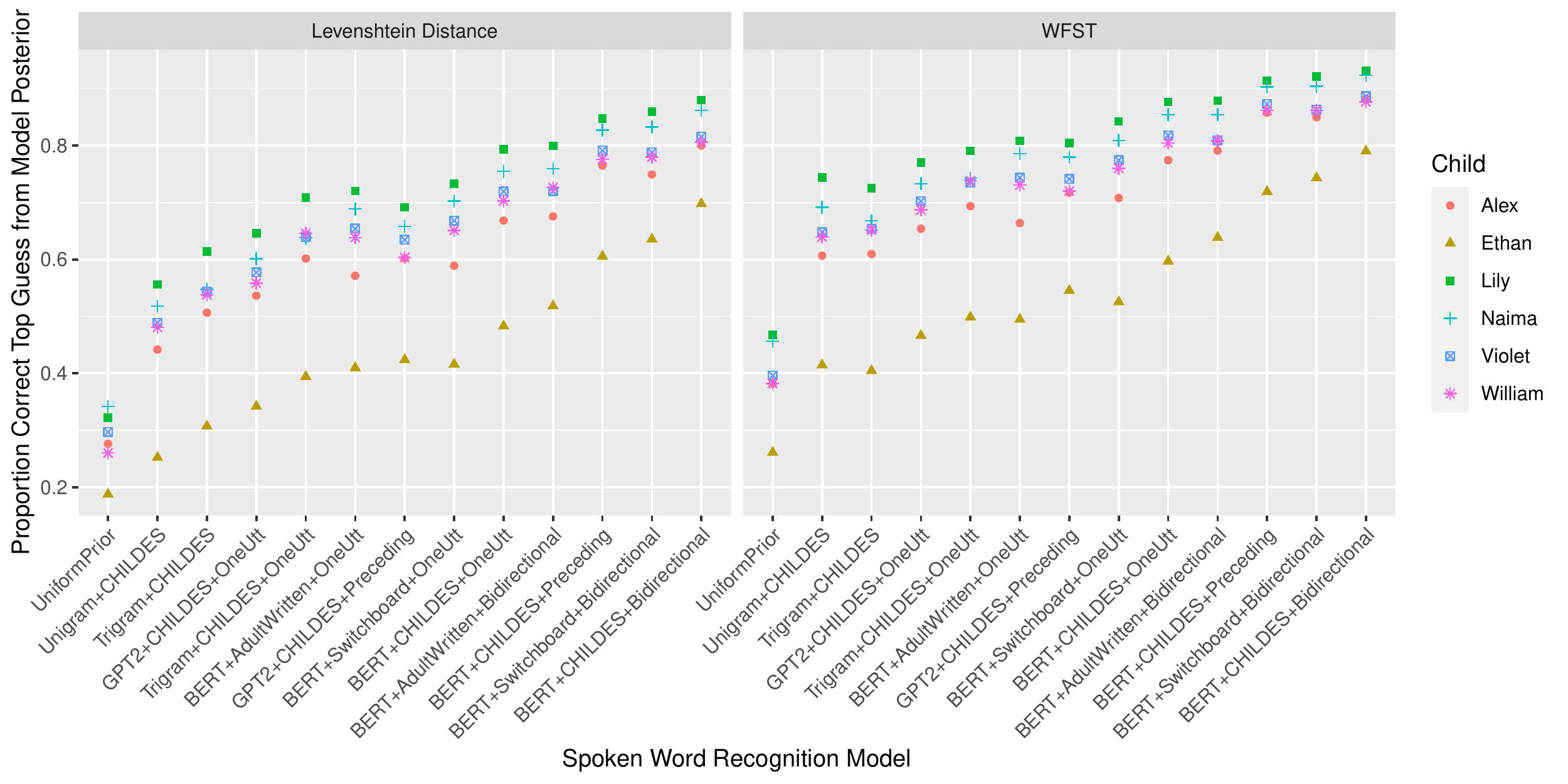}
\caption{
Proportion of test items where each model's highest posterior probability guess is the same the annotator's word identity, plotted by child across the thirteen models used in Expts. 1 and 2 (analogous to cross-child model results in ``Phoneme-Specific / Edit-Distance Correct Top Guess'' in Table 2). 
Results are faceted by the likelihood used by the model (Edit-Distance vs. Phoneme-Specific).
Models are plotted left to right based on average performance across children.
Model posteriors show a much more stable performance ordering across children than model priors (compare with Fig. S\ref{fig:prior_by_child}). 
}
\label{fig:posterior_by_child}
\end{figure*}

\end{document}